\documentclass[lettersize,journal]{IEEEtran}
\usepackage{amsmath,amsfonts}
\usepackage{algorithmic}
\usepackage{algorithm}
\usepackage{array}
\usepackage[caption=false,font=normalsize,labelfont=sf,textfont=sf]{subfig}
\usepackage{textcomp}
\usepackage{stfloats}
\usepackage{url}
\usepackage{verbatim}
\usepackage{graphicx}
\usepackage{cite}
\usepackage{threeparttable}
\usepackage[
    colorlinks=true,
    linkcolor=blue,
    citecolor=blue,
    urlcolor=black
]{hyperref}
\usepackage{etoolbox}
\AtBeginEnvironment{threeparttable}{\hypersetup{linkcolor=black, citecolor=black, urlcolor=black}} 
\AtBeginEnvironment{algorithm}{\hypersetup{linkcolor=black, citecolor=black, urlcolor=black}} 

\begin{document}

\title{LP-ICP: General Localizability-Aware Point Cloud Registration for Robust Localization in Extreme Unstructured Environments}

\author{Haosong Yue,~\IEEEmembership{Member,~IEEE,}
        Qingyuan Xu,
        Fei Chen,~\IEEEmembership{Senior Member,~IEEE,}
        Jia Pan,~\IEEEmembership{Senior Member,~IEEE,}
        and Weihai Chen,~\IEEEmembership{Member,~IEEE}}



\maketitle

\begin{abstract}
The Iterative Closest Point (ICP) algorithm is a crucial component of LiDAR-based SLAM algorithms. However, its performance can be negatively affected in unstructured environments that lack features and geometric structures, leading to low accuracy and poor robustness in localization and mapping. It is known that degeneracy caused by the lack of geometric constraints can lead to errors in 6-DOF pose estimation along ill-conditioned directions. Therefore, there is a need for a broader and more fine-grained degeneracy detection and handling method. This paper proposes a new point cloud registration framework, LP-ICP, that combines point-to-line and point-to-plane distance metrics in the ICP algorithm, with localizability detection and handling. Rather than relying solely on point-to-plane localizability information, LP-ICP enhances the localizability analysis by incorporating a point-to-line metric, thereby exploiting richer geometric constraints. It consists of a localizability detection module and an optimization module. The localizability detection module performs localizability analysis by utilizing the correspondences between edge points (with low local smoothness) to lines and planar points (with high local smoothness) to planes between the scan and the map. The localizability contribution of individual correspondence constraints can be applied to a broader range. The optimization module adds additional soft and hard constraints to the optimization equations based on the localizability category. This allows the pose to be constrained along ill-conditioned directions. The proposed method is evaluated on simulation and real-world datasets, showing comparable or better accuracy than the state-of-the-art methods in tested scenarios. Observed variations in partially localizable directions suggest the need for further investigation on robustness and generalizability.
\end{abstract}

\def\abstractname{Note to Practitioners}
\begin{abstract}
This paper was motivated by addressing the challenges of Simultaneous Localization and Mapping (SLAM) that use LiDAR as one of the sensors in extreme unstructured environments such as  planetary-like environments and underground tunnels. Due to the lack of features and geometric structures in these environments, the performance of pose estimation based on the Iterative Closest Point (ICP) algorithm is limited in the degenerate directions. This leads to low accuracy and poor robustness in localization and mapping. Most existing degeneracy detection methods have limited applicability or are not fine-grained. This paper proposes a new point cloud registration algorithm that additionally analyzes the localizability of point-to-line correspondences to detect and handle degeneracy in pose estimation. The proposed localizability detection method has the potential to be extended to other variants of ICP or multi-sensor fusion frameworks. The optimization module incorporates both soft and hard constraints based on the localizability analysis. Extensive experiments demonstrate that the proposed algorithm effectively detects localizability and achieves better or comparable performance to the state-of-the-art methods in challenging environments.
\end{abstract}


\section{Introduction}
\IEEEPARstart{S}{imultaneous} Localization and Mapping (SLAM) provides robots with pose information and a map of the surrounding environment, serving as a key technology in robotics. Currently, sensors such as LiDAR, cameras, and inertial measurement units (IMUs) are frequently used in SLAM systems. The point clouds from LiDAR can provide accurate distance measurements. Therefore, LiDAR-based methods have become an important branch of SLAM \cite{Cadena2016,zhang2014loam,Choi2023}.

Point cloud registration is a crucial step in LiDAR-based SLAM approaches. It aligns two sets of point clouds by estimating the rotation and translation between them \cite{biber2003normal,Min2020}. The Iterative Closest Point (ICP) algorithm is commonly used for scan-to-scan or scan-to-map registration to estimate the robot's pose. Common distance metrics include point-to-point \cite{Besl1992_p2po}, point-to-line \cite{Censi2008_p2l}, and point-to-plane \cite{low2004linear_p2pl}, \cite{Chen1992_p2pl}. These methods have been widely adopted in many advanced SLAM systems \cite{zhang2014loam, Lin2020_livox, Shan2021_lvisam, Xu2022_fastlio2, Sun2024_scelio}. In recent years, the ICP algorithm has continued to undergo further research \cite{Vizzo2023_kissicp, Ferrari2024_madicp, He2023_gfoicp}.

\begin{figure}[!t]
\includegraphics[ width=\linewidth]{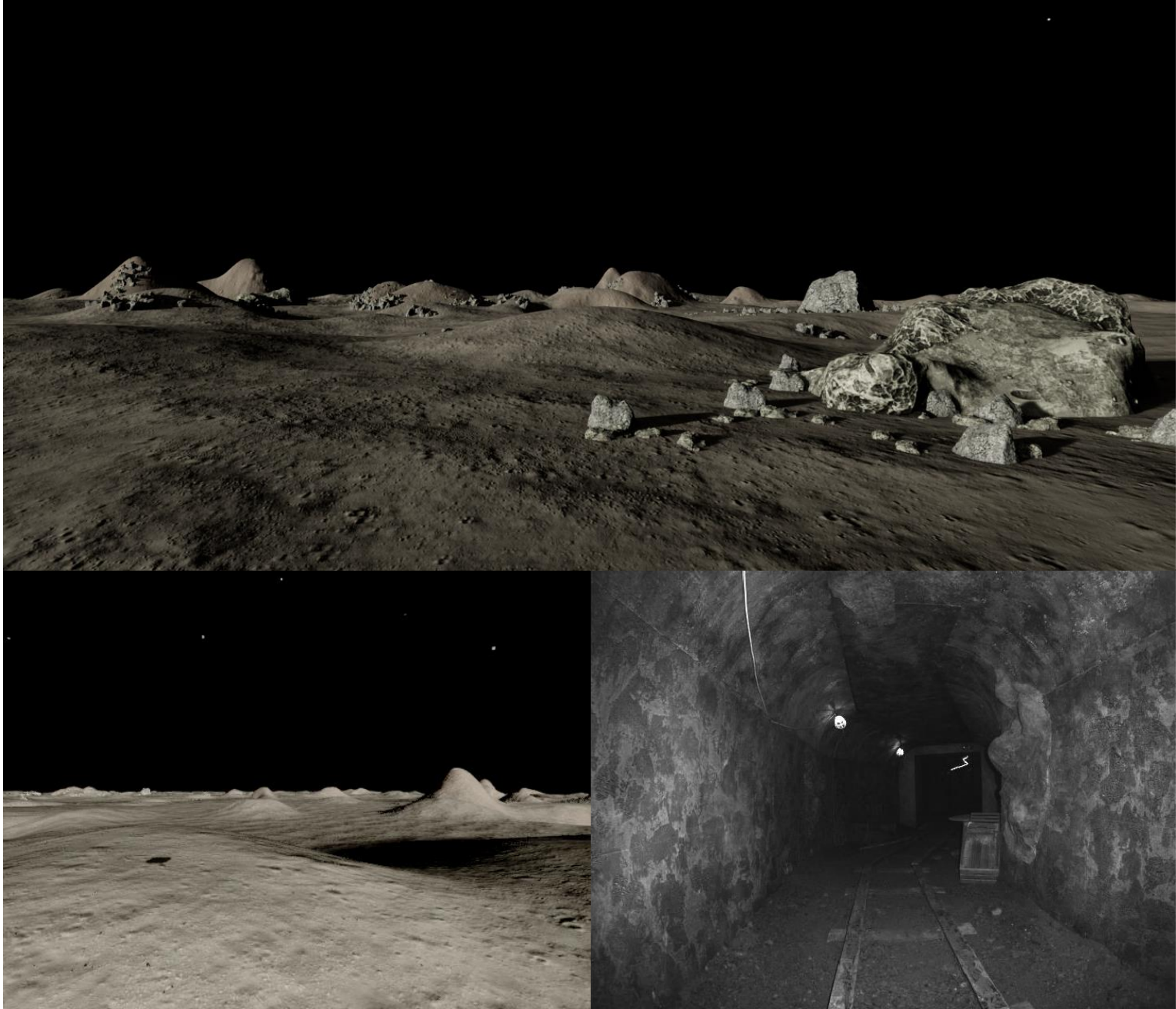}
\caption{Typical Degraded Environments: Planetary-like environment from our PLAM dataset and underground tunnel environment from the CERBERUS DARPA Subterranean Challenge Datasets \cite{Tranzatto2024_cerberus}.}
\label{fig:intro_environment}
\vspace{-1.2 em}
\end{figure}

 In geometrically rich environments, such as urban streets or campus classrooms, there are generally sufficient geometric constraints for existing point cloud registration methods. Therefore, the localization accuracy in these environments is quite satisfactory. However, extreme unstructured environments, such as planetary-like environments \cite{Giubilato2022_dlrdataset, LeGentil2020_gaussianLoop} on Mars or the Moon, and underground tunnels \cite{Tranzatto2024_cerberus, chen2024_hetero}, as shown in Fig.~\ref{fig:intro_environment}, as well as vast open areas, are characterized by self-similarity and sparse geometrical structures. This results in insufficient constraints being provided for ICP optimization. The optimization is rendered degenerate in certain directions of the 6-DoF pose, leading to decrease in the accuracy of localization and mapping.

\begin{figure*}[t]
\centering
\includegraphics[width=1.0\linewidth]{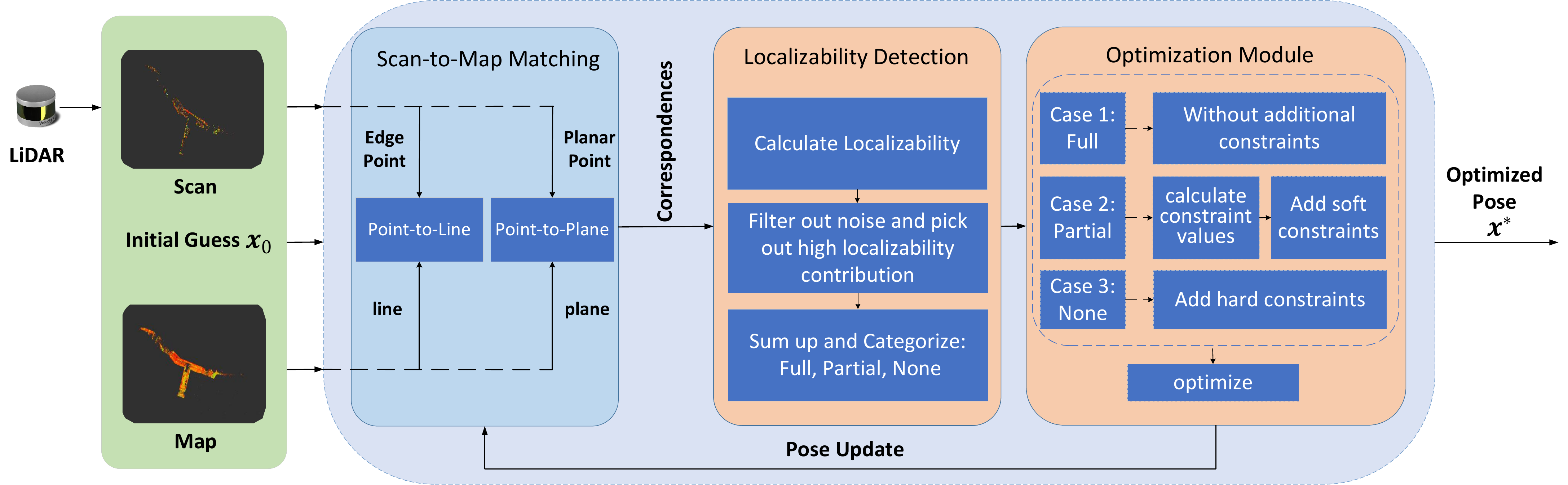}
\caption{Overview of the Proposed Point Cloud Registration Framework. The current scan from the LiDAR frame is transformed into the map frame using the initial estimate \(\boldsymbol{x}_0\) and enters the ICP loop together with the existing map. Next, LP-ICP, which combines point-to-line and point-to-plane distance metrics into the ICP framework, applies the proposed localizability detection and optimization modules to estimate the optimal pose.}
\label{fig:framework}
\end{figure*}

Different geometric constraint information, such as planes in various orientations or their edges, has varying impacts on the solution of ICP optimization. The evaluation of such impacts and the handling of different levels of degeneracy should be incorporated into the framework to improve the nature of the solution. Zhang et al. \cite{Zhang2016_zhang_etal} were the first to introduce the degeneracy factor into the optimization problem of state estimation, which can be used to detect the overall degeneracy level in each direction of the 6-DoF pose. They also employed the solution remapping after the binary judgment of degeneracy. However, it can only assess the overall degeneracy in a given direction and cannot be used to calculate the localizability contribution of individual constraints within it. Tuna et al. \cite{Tuna2024_xicp} recently introduced localizability detection of individual constraints in point-to-plane ICP for point cloud registration and established a ternary judgment for degeneracy, referred to as X-ICP. However, their method does not utilize point-to-line localizability information, which limits the potential for further accuracy improvement. Moreover, the formula for calculating the localizability contribution of individual constraints cannot be generalized to broader ICP algorithms and their variants. 

In this paper, we propose a new localizability-aware point cloud registration framework, LP-ICP, to address the LiDAR degeneracy problem. It is designed to improve the localization and mapping accuracy of SLAM systems in extreme unstructured environments. The proposed framework is illustrated in Fig.~\ref{fig:framework}. More specifically, LP-ICP is a point cloud registration framework that combines point-to-line and point-to-plane distance metrics in the ICP algorithm, with localizability detection and handling. The calculation formulas for the localizability contribution of individual correspondence constraints can be applied to other variants of ICP, or more generally, to pose estimation problems optimized using the Gauss-Newton method, such as estimating poses by minimizing the reprojection error in vision-based methods. Therefore, it has the potential to be extended to multi-sensor fusion frameworks. It consists of two modules: the localizability detection module (LocDetect-Module) and the optimization module with soft and hard constraints (Optimization-Module). During the scan-to-map registration, the LocDetect-module utilizes the correspondences between edge points and lines, as well as between planar points and planes. It analyzes the contribution strength along the principal directions of the optimization in the eigenspace. Then, we filter out the noise and pick out high localizability contributions. Further, the localizability of each direction of the 6-DoF pose is evaluated through a ternary threshold judgment, classified as: fully localizable, partially localizable, and non-localizable. 

The Optimization-Module first calculates additional constraints, utilizing localizability information from geometric constraints and correspondences with high localizability contributions. Then, based on the localizability categories from the LocDetect-Module, soft and hard constraints are added to the optimization equation. This ensures that the pose in non-localizable directions is not updated, maintaining the initial estimate unchanged. In partially localizable directions, the pose is updated under additional constraints. For fully localizable directions, the pose update remains unaffected. 

Additionally, to test the algorithm's performance in typical extreme unstructured environments, such as planetary-like environments, we collected a simulated SLAM dataset with LiDAR, IMU, and camera data from a visually realistic lunar-like environment. The dataset contains 10 sequences. The total trajectory length is 10 km. It can be used to test various components of the SLAM framework and different tasks in planetary exploration.

The proposed framework has been tested in multiple experiments on challenging simulation and real-world datasets. The experiments suggest that LP-ICP performs well in extreme unstructured environments and shows potential to improve the robot's localization capability and mapping accuracy in degraded environments. Compared to the state-of-the-art approaches \cite{Zhang2016_zhang_etal}, \cite{Tuna2024_xicp}, the proposed framework achieves comparable or improved accuracy in our tested scenarios.

To summarize, the contributions of this work are listed as follows:

\begin{itemize}
    \item We propose a new ICP algorithm framework with localizability detection and handling, utilizing point-to-line and point-to-plane localizability information. It introduces a novel point-to-line localizability analysis to help detect and handle degeneracy in extreme unstructured environments.
    \item We propose a general method for calculating the localizability contribution of individual correspondences. It can be used for pose estimation problems optimized using the Gauss-Newton method.
    \item A combined soft and hard constraint ICP optimization module, which utilizes localizability information, is developed to handle degeneracy.
    \item  We collected a multi-sensor simulated SLAM dataset from a visually realistic lunar-like environment, used for SLAM testing in typical extreme degraded environments.
    \item   A variety of experiments demonstrate the effectiveness of the proposed framework under our experimental conditions, in comparison with state-of-the-art methods.
\end{itemize}

\section{Related Work}
In this section, we first review the ICP method and its variants. After that, we survey studies related to degeneracy detection and handling.

\subsection{ICP Algorithm and Its Variants}
Point cloud registration aligns two or more sets of point clouds into a unified coordinate system by solving for the pose transformation between them. ICP is a commonly used method for point cloud registration. The ICP algorithm iteratively solves for the pose transformation between the source and target point clouds by minimizing a cost function, obtaining the optimal solution. Over the years, the ICP algorithm has evolved to include cost functions with different metrics, such as point-to-point \cite{Besl1992_p2po}, point-to-line \cite{Censi2008_p2l}, and point-to-plane \cite{low2004linear_p2pl}, \cite{Chen1992_p2pl}. Segal et al. \cite{Segal2010_geneicp} combined point-to-point and point-to-plane ICP into a probabilistic framework. Furthermore, Billings et al. \cite{Billings2015_imlp} introduced a noise model for further extension. Appropriately combining different error metrics can enhance the robustness of the algorithm across varying environments \cite{Lee2025_genzicp}. Lee et al. proposed GenZ-ICP \cite{Lee2025_genzicp}, which adaptively sets the weights of point-to-point and point-to-plane error metrics in the optimization problem based on the ratio of planar points to non-planar points. This mitigates the negative impact of optimization degeneracy in point-to-plane ICP in degraded environments. Serafin and Grisetti \cite{Serafin2015_nicp} proposed NICP, which incorporates angular errors of normal vectors into the cost function. Li et al. \cite{Li2024_adaptPR} considered both the angles and distances of local normals while introducing a point-to-surface version of ICP. Additionally, there are learning-based methods, such as \cite{Wang2019_deepclosestpoint}. Currently, the ICP algorithm combining point-to-line and point-to-plane distance metrics by utilizing edge and planar features in the environment remains one of the preferred choices. It has been applied in many advanced SLAM systems, such as \cite{zhang2014loam, Lin2020_livox, Shan2021_lvisam, Zhou2022_PLCLiSLAM}.

Currently, existing algorithms already demonstrate strong performance in feature-rich environments. However, in extreme unstructured environments, such as planetary-like environments or underground tunnels, they often fail due to the lack of features and structures. The lack of geometric constraints may lead to degeneracy in certain directions of 6-DoF pose estimation. When combined with the impact of noise, this can potentially result in an incorrect global optimal solution. Therefore, detecting and handling degeneracy is essential to obtain reliable pose estimation.

\subsection{Degeneracy Detection and Handling}
In recent years, many studies on degeneracy detection and handling have emerged. Gelfand et al. \cite{Gelfand_geometric_sample} proposed a point selection strategy to improve the geometric stability of the point-to-plane ICP algorithm. When selecting sampling points, the condition number of the covariance matrix is used as a stability measure. By increasing the sampling points in regions with fewer geometric constraints, the ICP algorithm reduces unstable transformations caused by sliding and improves accuracy. For point selection, the normal vector information of point-to-plane correspondences is utilized. However, it cannot handle situations where there are insufficient points in degenerate directions to allow for perfect optimization convergence. Shi et al. \cite{Shi2023_tim} proposed a degeneration-aware correlative scan matching (CSM) algorithm, DN-CSM, for 2D LiDAR. The algorithm extracts dense normal vector features from the current point cloud. By modeling and aggregating these features, it calculates the degeneracy direction and degree. It then utilizes a motion model with dynamic weighting to account for the degeneracy and implement CSM.

Zhang et al. \cite{Zhang2016_zhang_etal} pioneered the concept of the degeneracy factor, using the eigenvalues of the Hessian matrix to evaluate the overall degeneracy in the six directions of pose estimation. By solution remapping, they ensure that poses in degenerate directions do not update, thereby mitigating the negative impact of degeneracy. This method has been adopted in several LiDAR-based SLAM frameworks, such as \cite{zhang2014loam, Shan2021_lvisam, Shan2018_legoloam, Shan2020_liosam, Zhang2018_multi_slam, Jiao2022_multiLiDAR}. Hinduja et al. \cite{Hinduja2019} used the condition number to set a threshold for determining degeneracy in the point-to-plane ICP algorithm. During optimization, they applied the solution remapping technique. The results were incorporated into the SLAM pose graph optimization as partial loop closure factors, ensuring that ICP and pose graph updates and optimization are performed only in well-constrained directions of the state space. Lee et al. \cite{Lee2024_switchslam}, building on \cite{Zhang2016_zhang_etal}, utilized a chi-square distribution to non-heuristically set thresholds for the three most degenerate directions in LiDAR odometry. McDermott and Rife \cite{McDermott2022} proposed ICET, which enhances the Normal Distributions Transform (NDT) by introducing a condition number framework to detect global degeneracy and suppress solutions along degenerate directions while preserving information in useful directions.  This helps reduce geometric ambiguity along large flat surfaces. Wen et al. \cite{Wen2024_liver} set a threshold for binary detection of degeneracy by using the total residuals of the ICP algorithm in LiDAR odometry and the pose estimation results of the Visual-Inertial Odometry (VIO). In the degenerate direction of the LiDAR odometry, only visual measurements and IMU residuals are used for estimation. However, these methods are not fine-grained enough and cannot be used to evaluate the localizability contribution of individual constraints.

Zhen et al. \cite{Zhen2017_localiza2017} proposed a method that constructs an information matrix using the plane normal vectors from an existing map. They then use its eigenvalues to assess the localizability in the translational directions. However, they did not provide a detailed explanation or handling approach for the metrics. Furthermore, Zhen and Scherer \cite{Zhen2019_localiza2019} in the point-to-plane ICP algorithm used the sensitivity of LiDAR measurements w.r.t. the 6-DoF pose as a measure of localizability to assess the strength of the constraints.  They also utilized the geometric information from the point and plane normals. To handle low localizability, they chose to incorporate measurements from other sensors for compensation. Tuna et al. \cite{Tuna2024_xicp} recently introduced localizability detection into the point-to-plane ICP algorithm, utilizing point and plane normal information, referred to as X-ICP. They pioneered a ternary degeneracy judgment mechanism for each direction of the pose solution, thereby incorporating more constraint information. The localizability calculation formula for individual constraints is explained from a geometric or force perspective. However, further mathematical derivation and explanation from the perspective of degeneracy are necessary. Chen et al. \cite{Chen2024_relead} adopted the degeneracy detection method proposed in \cite{Tuna2024_xicp} for LiDAR-inertial odometry and incorporated the computed additional constraints into the ESIKF update. Tuna et al. \cite{tuna2025_informed} proposed and introduced new methods such as inequality constraints, truncated SVD, and Tikhonov regularization into the field of degeneracy-aware point cloud registration. They also evaluated and analyzed the effectiveness of these different constraint types in comparison with existing methods. Hatleskog and Alexis \cite{Hatleskog2024_prob_degeDec} proposed a method in the point-to-plane ICP algorithm that considers the noise in both point and plane normals. By accounting for the probabilistic characteristics of the noise entering the Hessian, their approach helps to detect degeneracy and reduces the impact of noise in degenerate directions. These studies provide more fine-grained detection of localizability and enable the evaluation of the localizability contribution of individual constraints.  However, they are limited to the point-to-plane ICP algorithm and do not utilize the localizability information from point-to-line correspondences, which limits further improvements in accuracy.

Nubert et al. \cite{Nubert2022_learnLocliza} proposed using a learning-based approach to detect the localizability of LiDAR. The method is trained with simulated data and utilizes the point-to-plane ICP algorithm. Degeneracy detection does not require heuristic threshold adjustments, but it is only applicable to scan-to-scan point cloud registration. There are also learning-based methods such as \cite{Li2019_learn1, Gao2020_learn2}. However, their reliance on labeled ground truth data and high computational demands limits their use in resource-constrained systems.

\section{Problem Formulation, Preliminaries and System Overview}
This section introduces the preliminaries of point cloud registration and degeneracy detection, as well as the problem we aim to address. Finally, an overview of the LP-ICP system framework is provided.

\subsection{Optimization-Based Point Cloud Registration}
Nonlinear optimization plays a crucial role in many SLAM systems\cite{Cadena2016, Qin2018_vinsmono, zhang2014loam}. An optimization-based state estimation problem can be formulated by solving a function:

\begin{equation}
\label{equa1}
\arg\min_{\boldsymbol{x}} \boldsymbol{f}^2(\boldsymbol{x}).
\end{equation}

Here, \(\boldsymbol{x}\) is an \(n \times 1\) state vector, where \(n\) is the dimension of the state space.

In point cloud registration, the optimal solution can be obtained by iteratively solving for the pose transformation between the source and target point clouds through minimizing a cost function. Let \(f_i(\boldsymbol{x})\) denotes the residual for each correspondence between the source and target point clouds. The problem is then equivalent to solving the following:

\begin{equation}
\label{equa2}
\arg\min_{\boldsymbol{x} \in \mathbb{R}^6} \sum_{i=1}^N \| f_i(\boldsymbol{x}) \|^2.
\end{equation}

Here, \(\boldsymbol{x} = (\boldsymbol{r}^T\;\boldsymbol{t}^T)^T \in \mathbb{R}^6\). \(\boldsymbol{x}\) is the pose state vector, where \(\boldsymbol{r}\) denotes the 3-DoF rotation representation, \(\boldsymbol{t}\) denotes the 3-DoF translation vector, and \(N\) denotes the number of matching correspondences.

Currently, the ICP algorithm combining point-to-line and point-to-plane distance metrics remains one of the preferred choices. It has been applied in many advanced SLAM systems, such as \cite{zhang2014loam, Lin2020_livox, Shan2021_lvisam, Zhou2022_PLCLiSLAM}. To balance efficiency, accuracy, and robustness, representative geometric feature points are often extracted for optimization. For example, edge points are used with point-to-line cost functions, and planar points with point-to-plane cost functions, with both types of points combined to participate together in the ICP optimization \cite{zhang2014loam}.

The combined point-to-line and point-to-plane ICP minimization problem is defined as follows:

\begin{equation}
\label{equa3}
\min_{\boldsymbol{R},\boldsymbol{t}} \sum_{i=1}^{N_1} \| f_{ei}(\boldsymbol{x}) \|^2 + \sum_{j=1}^{N_2} \| f_{pj}(\boldsymbol{x}) \|^2
\end{equation}
where 
\begin{equation}
\label{equa4}
f_{ei}(\boldsymbol{x}) = \| (\boldsymbol{R} \boldsymbol{p}_i^L + \boldsymbol{t} - \boldsymbol{q}_i^M) \times \boldsymbol{l}_i^M \|
\end{equation}

\begin{equation}
\label{equa5}
f_{pj}(\boldsymbol{x}) = (\boldsymbol{R} \boldsymbol{p}_j^L + \boldsymbol{t} - \boldsymbol{q}_j^M) \cdot \boldsymbol{n}_j^M .
\end{equation}

Here, \(f_{ei}(\boldsymbol{x})\) is the residual for an edge point to a line. \(f_{pj}(\boldsymbol{x})\) is the residual for a planar point to a plane. \(N_1\) is the number of edge points, \(N_2\) is the number of planar points, and the total number of points is \(N=N_1+N_2\). 

Specifically, the alignment of the current scan point cloud \(\boldsymbol{p}^L\) in the LiDAR frame (denoted as \(L\)) with the map point cloud \(\boldsymbol{q}^M\) in the map frame (denoted as \(M\)) is achieved through the transformation \(\boldsymbol{T} = \begin{bmatrix} \boldsymbol{R} \;|\; \boldsymbol{t} \end{bmatrix}\) (or  \(\boldsymbol{x} = (\boldsymbol{r}^T\;\boldsymbol{t}^T)^T \in \mathbb{R}^6\)). Here, \(\boldsymbol{R} \in SO(3)\) denotes the rotation matrix and  \(\boldsymbol{t} \in \mathbb{R}^3\) denotes the translation vector. \(\boldsymbol{p}_i^L\) and \(\boldsymbol{p}_j^L\) are the edge and planar points extracted from the LiDAR frame, respectively. \(\boldsymbol{q}_i^M\) and \(\boldsymbol{q}_j^M\) are the corresponding matching points in the map frame, typically found using a search method, such as k-d tree search. \(\boldsymbol{l}_j^M\) is the unit direction vector of the edge line passing through \(\boldsymbol{q}_i^M\), and \(\boldsymbol{n}_j^M\) is the unit normal vector of the plane passing through \(\boldsymbol{q}_j^M\).

The problem can be solved iteratively using the Gauss-Newton method.

\subsection{Degeneracy Detection}
In degraded environments, the performance of point cloud registration deteriorates. Therefore, it is necessary to detect and handle degeneracy. Zhang et al. \cite{Zhang2016_zhang_etal} pioneered the use of a degeneracy factor \(D\) to evaluate whether the 6-DoF pose solution \(\boldsymbol{x}\) is degenerate in a certain direction. The meaning of \(D\) is expressed as follows: by perturbing the constraint with a displacement of distance \(\delta d\), the resulting shift \(\delta x_c\) of the true solution \(\boldsymbol{x}_0\) in the direction of perturbation \(\boldsymbol{c}\) is given by  \(D = \frac{\delta d}{\delta x_c}\). In other words, \(D\) denotes the stiffness of the solution under constraint perturbation. If \(D\) is smaller, meaning that for the same perturbation \(\delta d\), the amount of movement \(\delta x_c\) of the solution in that direction is larger. It indicates that the stiffness w.r.t. disturbances is lower in that direction. Through the mathematical derivation in \cite{Zhang2016_zhang_etal}, it is concluded that:

\begin{equation}
\label{equa6}
D = \frac{\delta d}{\delta x_c} = \lambda + 1.
\end{equation}

Here, \(\lambda\) is the eigenvalue of the matrix \(\boldsymbol{A}^T \boldsymbol{A}\), which can be computed using eigenvalue decomposition. Note that the eigenvalues of \(\boldsymbol{A}^T \boldsymbol{A}\) are typically referred to as singular values and computed using the singular value decomposition. \(\boldsymbol{A}\) comes from the following equation (i.e. the linearized form of \eqref{equa1}):

\begin{equation*}
\label{equa7}
\arg\min_{\boldsymbol{x}} \| \boldsymbol{A}\boldsymbol{x} - \boldsymbol{b} \|.
\end{equation*}

The Gauss-Newton method is a commonly used approach for solving optimization problems. In optimization problems solved using the Gauss-Newton method, \(\lambda\) is the eigenvalues corresponding to the Hessian matrix of the original optimization problem. \(\lambda\) can be used to evaluate whether optimization in a specific direction suffers from degeneracy. However, it cannot assess the localizability contribution of individual correspondence constraints in that direction. Therefore, obtaining more fine-grained localizability detection results is one of the objectives of this study.

\subsection{System Overview}
LP-ICP is divided into two modules: the Localizability Detection Module (LocDetect-Module) and the Optimization Module with soft and hard constraints (Optimization-Module). The proposed framework is shown in Fig.~\ref{fig:framework}. LP-ICP is integrated into the point cloud registration framework of LVI-SAM \cite{Shan2021_lvisam} for evaluation and validation. LVI-SAM integrates the LIO (LiDAR-Inertial Odometry) submodule and the VIO (Visual-Inertial Odometry) submodule. The LIO submodule is the primary component, and the VIO submodule provides the initial estimates to the LIO submodule. Note that in LVI-SAM, the 3-DOF rotation in the pose are represented using Euler angles.

The LocDetect-Module first derives the formula for calculating the localizability contribution of individual correspondence constraints. Note that in the ICP algorithm, we refer to a correspondence used to constrain pose estimation, such as a point-to-line or point-to-plane correspondence, as an individual correspondence constraint. This approach is not only applicable to point-to-line and point-to-plane ICP algorithms but can also be extended to pose estimation problems optimized using the Gauss-Newton method, such as variants of ICP or pose estimation through optimizing the reprojection error. This corresponds to Section~\ref{section:detection_module_A}. Then, the localizability contribution vectors for each edge point \(\boldsymbol{p}_i\) in the point-to-line metric are computed:  \(\boldsymbol{F}_{ri}^e \in \mathbb{R}^{3 \times 1}\) in the rotation direction and \(\boldsymbol{F}_{ti}^e \in \mathbb{R}^{3 \times 1}\) in the translation direction. Similarly, the localizability contribution vectors for each planar point \(\boldsymbol{p}_j\) in the point-to-plane metric are computed: \(\boldsymbol{F}_{rj}^p \in \mathbb{R}^{3 \times 1}\) in the rotation direction and \(\boldsymbol{F}_{tj}^p \in \mathbb{R}^{3 \times 1}\) in the translation direction. The localizability vectors in each direction of the 6-DOF pose are aggregated to assess the localizability in the corresponding direction. They are then compared with thresholds to categorize the localizability into three types: non-localizable (None), partially localizable (Partial), and fully localizable (Full). This corresponds to Section~\ref{section:detection_module_B} and Section~\ref{section:detection_module_C}.

The Optimization-Module (Section~\ref{section:optimization_module}) utilizes the categorization results from the LocDetect-Module for each direction of the 6-DOF pose, along with higher localizability contributions. It constructs optimization equations to solve for the optimal pose estimate \(\boldsymbol{x}^*\). We use a method combining soft and hard constraints. Hard constraints \cite{Tuna2024_xicp} are applied in non-localizable directions to prevent pose updates in those directions. Soft constraints are added in partially localizable directions, allowing the pose to be updated under the constraint. The pose update in fully localizable directions is unaffected.

\section{Localizability Detection Module}
\label{section:detection_module}
This section first derives the formula for calculating the localizability contribution of individual correspondence constraints, and then elaborates on the LocDetect-Module.

\subsection{Localizability of an Individual Correspondence Constraint}
\label{section:detection_module_A}
It is known that the degeneracy factor \(D\) can be used to evaluate the level of degeneracy in each direction of the 6-DOF pose, and it is applicable to general pose estimation problems.

If the pose estimation problem in \eqref{equa2} is solved using the Gauss-Newton method, then \(\lambda\) in \eqref{equa6} is also the eigenvalue corresponding to the Hessian matrix of the original optimization problem in \eqref{equa2}. The Hessian matrix is as follows:

\begin{equation}
\label{equa8}
\boldsymbol{H} = \boldsymbol{J}^T \boldsymbol{J} = \sum_{i=1}^{N} \boldsymbol{J}_i^T \boldsymbol{J}_i \in \mathbb{R}^{6 \times 6}
\end{equation}
where \(\boldsymbol{J}_i\) is the Jacobian matrix of \({f}_i(\boldsymbol{x})\) w.r.t. the pose \(\boldsymbol{x} = (\boldsymbol{r}^T\;\boldsymbol{t}^T)^T \in \mathbb{R}^6\), i.e. \(\boldsymbol{J}_i = \frac{\partial f_i(\boldsymbol{x})}{\partial \boldsymbol{x}} = ( \frac{\partial f_i(\boldsymbol{x})}{\partial x_1}, \ldots, \frac{\partial f_i(\boldsymbol{x})}{\partial x_6} ) \in \mathbb{R}^{1 \times 6}\). So for \(N\) correspondence constraints, we have: \(\boldsymbol{J}=(\boldsymbol{J}_1^T \cdots \boldsymbol{J}_N^T)^T \in \mathbb{R}^{N \times 6}\).

Since \(\boldsymbol{H}\) in \eqref{equa8} is a real symmetric positive definite matrix, it can be diagonalized by eigenvalue decomposition as follows:

\begin{equation*}
\label{equaH_VAV}
\boldsymbol{H} = \boldsymbol{V} \boldsymbol{\Lambda} \boldsymbol{V}^T.
\end{equation*}

Here, \(\boldsymbol{V} = (\boldsymbol{v}_1 \quad \cdots \quad \boldsymbol{v}_6)\) are the eigenvectors in matrix form and \(\boldsymbol{v}_i \in \mathbb{R}^{6 \times 1}\) is the eigenvector corresponding to the eigenvalues \(\lambda_i\) in the diagonal matrix \(\boldsymbol{\Lambda}\).

So we obtain:

\begin{equation*}
\label{equaA_VHV}
\boldsymbol{\Lambda} = \boldsymbol{V}^T \boldsymbol{H} \boldsymbol{V} = \boldsymbol{V}^T \boldsymbol{J}^T \boldsymbol{J} \boldsymbol{V} = \boldsymbol{(JV)}^T \boldsymbol{JV}.
\end{equation*}

Here, the calculation formula for \(\boldsymbol{J} \boldsymbol{V}\) is as follows:

\begin{equation*}
\label{equaJV_MAT}
\boldsymbol{JV} = 
\begin{pmatrix}
\boldsymbol{J}_1 \boldsymbol{v}_1 & \cdots & \boldsymbol{J}_1 \boldsymbol{v}_6 \\
\vdots & \ddots & \vdots \\
\boldsymbol{J}_N \boldsymbol{v}_1 & \cdots & \boldsymbol{J}_N \boldsymbol{v}_6
\end{pmatrix}
\in \mathbb{R}^{N \times 6}
\end{equation*}
where \(\boldsymbol{J}_i \boldsymbol{v}_j\) denotes the projection of  \(\boldsymbol{J}_i\) in the direction of the eigenvector \(\boldsymbol{v}_j\).

Since \(\Lambda\) is a diagonal matrix, we have:

\begin{equation*}
\label{equa9}
\begin{aligned}
\boldsymbol{\Lambda} &= (\boldsymbol{J} \boldsymbol{V})^T \boldsymbol{J} \boldsymbol{V} \\
&= \begin{pmatrix}
\sum_{i=1}^{N} (\boldsymbol{J}_i \boldsymbol{v}_1)^2 & 0  & 0 \\
0 & \ddots  & 0 \\
0 & 0 & \sum_{i=1}^{N} (\boldsymbol{J}_i \boldsymbol{v}_6)^2
\end{pmatrix} \in \mathbb{R}^{6 \times 6} \\
&= \begin{pmatrix}
\lambda_1 & 0  & 0 \\
0 & \ddots & 0 \\
0 & 0 & \lambda_6
\end{pmatrix}.
\end{aligned}
\end{equation*}

Thus, we obtain:

\begin{equation}
\label{equa10}
\lambda_j = \sum_{i=1}^{N} (\boldsymbol{J}_i \boldsymbol{v}_j)^2
\end{equation}

That is, \(\lambda_j\) equals the sum of the squared projections of \(\boldsymbol{J}_i\) onto the eigenvector \(\boldsymbol{v}_j\), where  \(\boldsymbol{J}_i\) is the Jacobian of the i-th residual \(f_i(\boldsymbol{x})\) w.r.t. the pose \(\boldsymbol{x}\).

According to \eqref{equa6}, \(\lambda_j\) can be compared with a threshold to determine whether the state estimation problem in \eqref{equa2} is degenerate in the \(\boldsymbol{v}_j\) direction. \(\lambda_j\) can be used to evaluate the level of degeneracy in the \(\boldsymbol{v}_j\) direction. This is equivalent to using \eqref{equa10} to compare with the threshold to assess degeneracy in the \(\boldsymbol{v}_j\) direction. In other words, it evaluates the localizability of the \(N\) correspondence constraints in the \(\boldsymbol{v}_j\) direction.

We take one of the terms in the summation formula \eqref{equa10} and define \(F_{ji}\) as the localizability contribution of an individual correspondence constraint, as follows:

\begin{equation}
\label{equa11}
F_{ji} = (\boldsymbol{J}_i \boldsymbol{v}_j)^2
\end{equation}

It is used to evaluate the localizability contribution of the i-th correspondence constraint in the direction along \(\boldsymbol{v}_j\).

So we have:

\begin{equation}
\label{equa12}
\lambda_j = \sum_{i=1}^{N} F_{ji}
\end{equation}

Note that we provide two equations similar to \eqref{equa11} and \eqref{equa12}, as follows:

\begin{equation}
\label{equa13}
L_j = \sum_{i=1}^{N} \left| \boldsymbol{J}_i \boldsymbol{v}_j \right|
\end{equation}

\begin{equation}
\label{equa14}
F_{ji} = \left| \boldsymbol{J}_i \boldsymbol{v}_j \right|
\end{equation}

We consider \eqref{equa11} and \eqref{equa14}, as well as \eqref{equa12} and \eqref{equa13}, to be similar in nature but different in value. When \eqref{equa13} and \eqref{equa14} are used in the point-to-plane ICP algorithm, and rotation is represented using Lie algebra, they become identical to the localizability evaluation formula of X-ICP \cite{Tuna2024_xicp}.

For the convenience of subsequent calculations and usage, we choose to decouple translation and rotation in the localizability analysis in the above derivation. Tuna et al. \cite{Tuna2024_xicp} provides an explanation for performing localizability analysis separately in the translational and rotational eigenspaces from the perspective of information analysis. Note that we do not explicitly handle cases where the eigenvectors couple rotational and translational states. For \(N\) correspondence constraints, we have: \(\boldsymbol{J}_r = (\boldsymbol{J}_{r1}^T  \cdots \boldsymbol{J}_{rN}^T)^T \in \mathbb{R}^{N \times 3}\) in the rotation directions and \(\boldsymbol{J}_t =( \boldsymbol{J}_{t1}^T \cdots \boldsymbol{J}_{tN}^T)^T \in \mathbb{R}^{N \times 3}\) in the translation directions, where \(\boldsymbol{J}_{ri} \in \mathbb{R}^{1 \times 3}\) is the Jacobian of the i-th residual \(f_i \boldsymbol(x)\) w.r.t. the 3-DOF rotation and \(\boldsymbol{J}_{ti} \in \mathbb{R}^{1 \times 3}\) is the Jacobian of the i-th residual \(f_i \boldsymbol(x)\) w.r.t. the 3-DOF translation. That is, it is equivalent to performing eigenvalue decomposition on the top-left 3\(\times\)3 matrix \(\boldsymbol{H}_r\) and the bottom-right 3\(\times\)3 matrix \(\boldsymbol{H}_t\) of the Hessian matrix \(\boldsymbol{H}\), as follows:

\begin{equation}
\label{equa15}
\boldsymbol{H}_r = \boldsymbol{J}_r^T \boldsymbol{J}_r = \boldsymbol{V}_r \boldsymbol{\Lambda}_r \boldsymbol{V}_r^T
\end{equation}

\begin{equation}
\label{equa16}
\boldsymbol{H}_t = \boldsymbol{J}_t^T \boldsymbol{J}_t = \boldsymbol{V}_t \boldsymbol{\Lambda}_t \boldsymbol{V}_t^T
\end{equation}

In the rotation direction, we have \(\boldsymbol{V}_r=(\boldsymbol{v}_{r1} \; \boldsymbol{v}_{r2} \; \boldsymbol{v}_{r3})\), where \(\boldsymbol{v}_{rj} \in \mathbb{R}^{3 \times 1}\) is the eigenvector corresponding to the eigenvalue \(\lambda_j\) in the diagonal matrix \(\boldsymbol{\Lambda}_r\). In the translation direction, we have \(\boldsymbol{V}_t=(\boldsymbol{v}_{t1} \; \boldsymbol{v}_{t2} \; \boldsymbol{v}_{t3})\), where \(\boldsymbol{v}_{tj} \in \mathbb{R}^{3 \times 1}\) is the eigenvector corresponding to the eigenvalue \(\lambda_j\) in the diagonal matrix \(\boldsymbol{\Lambda}_t\).

Similarly, we can derive conclusions analogous to \eqref{equa11} and \eqref{equa12}. In the rotation direction, we obtain:

\begin{equation*}
\label{equa17}
\lambda_j = \sum_{i=1}^{N} (\boldsymbol{J}_{ri} \boldsymbol{v}_{rj})^2
\end{equation*}
where \(j=1,2,3\). That is, \(\lambda_j\) equals the sum of the squared projections of \(\boldsymbol{J}_{ri}\) onto the rotation-related eigenvector \(\boldsymbol{v}_{rj}\).

The localizability contribution vectors of the i-th correspondence constraint in the three directions \(\boldsymbol{v}_{r1}\), \(\boldsymbol{v}_{r2}\) and \(\boldsymbol{v}_{r3}\) of the rotational eigenspace are as follows: 

\begin{equation}
\label{equa18}
\boldsymbol{F}_{ri} = \begin{pmatrix}
(\boldsymbol{J}_{ri} \boldsymbol{v}_{r1})^2 \\
(\boldsymbol{J}_{ri} \boldsymbol{v}_{r2})^2 \\
(\boldsymbol{J}_{ri} \boldsymbol{v}_{r3})^2
\end{pmatrix} \in \mathbb{R}^{3 \times 1}
\end{equation}

Similarly, in the translation direction, we have:

\begin{equation*}
\label{equa19}
\lambda_j = \sum_{i=1}^{N} (\boldsymbol{J}_{ti} \boldsymbol{v}_{tj})^2
\end{equation*}
where \(j=1,2,3\).

The localizability contribution vectors of the i-th correspondence constraint in the three directions \(\boldsymbol{v}_{t1}\), \(\boldsymbol{v}_{t2}\) and \(\boldsymbol{v}_{t3}\) of the translational eigenspace are as follows: 

\begin{equation}
\label{equa20}
\boldsymbol{F}_{ti} = \begin{pmatrix}
(\boldsymbol{J}_{ti} \boldsymbol{v}_{t1})^2 \\
(\boldsymbol{J}_{ti} \boldsymbol{v}_{t2})^2 \\
(\boldsymbol{J}_{ti} \boldsymbol{v}_{t3})^2
\end{pmatrix} \in \mathbb{R}^{3 \times 1}
\end{equation}

Eq. \eqref{equa18} and \eqref{equa20} are used to evaluate the localizability contribution of the i-th correspondence constraint in each direction of the 6-DOF pose. And subsequent calculations are performed using \eqref{equa18} and \eqref{equa20}.

\subsection{Localizability Contribution of Edge Point-to-Line and Planar Points-to-Plane Correspondences}
\label{section:detection_module_B}
Here, we compute the localizability contribution of each edge point-to-line correspondence constraint and each planar point-to-plane correspondence constraint in \eqref{equa3}, respectively.

The localizability contribution vector for an individual correspondence constraint can be computed using \eqref{equa18} and \eqref{equa20}. First, we calculate the Jacobian, and then project it onto the eigenspace to obtain the localizability contribution vector for each correspondence constraint.

Note that for the 3-DOF rotation, we will provide two versions of the Jacobian for reference: one calculated using Lie algebra and the other using Euler angles. We integrate LP-ICP into the point cloud registration framework of the LVI-SAM \cite{Shan2021_lvisam}. Since the rotation in the LVI-SAM algorithm is represented using Euler angles, we will use the Euler angle-based Jacobian for subsequent calculations.

\subsubsection{Jacobian of the Point-to-Line Cost Function}
In the point-to-line cost function \eqref{equa4}, the Jacobians of the distance residual \(\boldsymbol{f}_{ei}\) w.r.t. translation \(\boldsymbol{t}\) for the edge point \(\boldsymbol{p}_i^L=(p_{ix}, \,p_{iy},\,p_{iz})^T \in \mathbb{R}^{3 \times 1}\) are calculated as follows:

\begin{equation}
\label{equa21}
\boldsymbol{J}_{ti}^e = (\boldsymbol{d}_i^M)^T.
\end{equation}

Here, \(\boldsymbol{d}_i^M=(d_{ix}, \,d_{iy},\,d_{iz})^T \in \mathbb{R}^{3 \times 1}\) is the unit distance vector from the edge point \(\boldsymbol{p}_i^M\) to the associated line in the map frame.

The Jacobians of the distance residual \({f}_{ei}(\boldsymbol{x})\) w.r.t. rotation \(\boldsymbol{r}\) for the edge point \(\boldsymbol{p}_i^L\) are calculated, when the rotation is represented using Lie algebra, as follows:

\begin{equation}
\label{equa22}
\boldsymbol{J}_{ri}^e = \left( \boldsymbol{p}_i^L \times \boldsymbol{d}_i^L \right)^T.
\end{equation}

When the rotation is represented using Euler angles \((\alpha, \,\beta, \, \gamma)\), it is as follows:

\begin{equation} 
\label{equa23}
\boldsymbol{J}_{ri}^e = \left( \boldsymbol{J}_{rix}^e,\, \boldsymbol{J}_{riy}^e ,\, \boldsymbol{J}_{riz}^e \right)
\end{equation}

\begin{equation*} 
\label{equq23_components}
\begin{aligned}
&\text{where}\\ 
&\boldsymbol{J}_{rix}^e = \left[ (s\alpha s\gamma + c\alpha s\beta c\gamma) p_{iy} + (s\alpha c\gamma - c\alpha s\beta s\gamma) p_{iz} \right] \times d_{ix}^M \\
&\quad + \left[ (-c\alpha s\gamma + s\alpha s\beta c\gamma) p_{iy} + (-c\alpha c\gamma - s\alpha s\beta s\gamma) p_{iz} \right] \times d_{iy}^M \\
&\quad + \left[ c\beta c\gamma p_{iy} - c\beta s\gamma p_{iz} \right] \times d_{iz}^M, \\
&\boldsymbol{J}_{riy}^e = \left[ -c\alpha s\beta p_{ix} + (c\alpha c\beta s\gamma) p_{iy} + (c\alpha c\beta c\gamma) p_{iz} \right] \times d_{ix}^M \\
&\quad + \left[ -s\alpha s\beta p_{ix} + (s\alpha c\beta s\gamma) p_{iy} + (s\alpha c\beta c\gamma) p_{iz} \right] \times d_{iy}^M \\
&\quad + \left[ -c\beta p_{ix} - s\beta s\gamma p_{iy} - s\beta c\gamma p_{iz} \right] \times d_{iz}^M, \\
&\boldsymbol{J}_{riz}^{e} = [ -s \alpha c \beta p_{ix} + (-c \alpha c \gamma - s \alpha s \beta s \gamma) p_{iy} \\
&\quad + (c \alpha s \gamma - s \alpha s \beta c \gamma) p_{iz} ] \times d_{ix}^{M} \\
&\quad + [ c \alpha c \beta p_{ix} + (-s \alpha c \gamma + c \alpha s \beta s \gamma) p_{iy} \\
&\quad + (s \alpha s \gamma + c \alpha s \beta c \gamma) p_{iz} ] \times d_{iy}^{M}.
\end{aligned}
\end{equation*}

Here, \(s\alpha=sin(\alpha)\), \(c\alpha=cos(\alpha)\), and similarly for others.

Subsequent calculations for \(\boldsymbol{J}_{ti}^e\) and \(\boldsymbol{J}_{ri}^e\) in this paper will use \eqref{equa21}, and \eqref{equa23}, respectively.

\subsubsection{Jacobian of the Point-to-Plane Cost Function}
Similarly, in the point-to-plane cost function \eqref{equa5}, the Jacobian of the distance residual \({f}_{pi}(\boldsymbol{x})\) w.r.t. the translation \(\boldsymbol{t}\) for the planar point \(\boldsymbol{p}_i^L = ( p_{ix},\, p_{iy},\, p_{iz})^T \in \mathbb{R}^{3 \times 1}\) is calculated as follows:

\begin{equation}
\label{equa24}
\boldsymbol{J}_{ti}^{p} = (\boldsymbol{n}_{i}^{M})^T.
\end{equation}

Here, \(\boldsymbol{n}_i^M = ( n_{ix},\, n_{iy},\, n_{iz})^T \in \mathbb{R}^{3 \times 1}\) is the unit normal vector of the plane associated with the planar point \(\boldsymbol{p}_i^M\) in the map frame.

The Jacobians of the distance residual \(f_{pi}(\boldsymbol{x})\) w.r.t. rotation \(\boldsymbol{r}\) for the planar point \(\boldsymbol{p}_i^L\) are calculated, when the rotation is represented using Lie algebra, as follows:

\begin{equation}
\label{equa25}
\boldsymbol{J}_{ri}^{p} = \left(\boldsymbol{p}_{i}^{L} \times \boldsymbol{n}_{i}^{L}\right)^T.
\end{equation}

When the rotation is represented using Euler angles \((\alpha, \,\beta, \, \gamma)\), it is as follows:

\begin{equation} 
\label{equa26}
\boldsymbol{J}_{ri}^p = \left( \boldsymbol{J}_{rix}^p,\, \boldsymbol{J}_{riy}^p ,\, \boldsymbol{J}_{riz}^p \right)
\end{equation}

\begin{equation*} 
\label{equq26_components}
\begin{aligned}
&\text{where}\\
&\boldsymbol{J}_{rix}^p = \left[ (s\alpha s\gamma + c\alpha s\beta c\gamma) p_{iy} + (s\alpha c\gamma - c\alpha s\beta s\gamma) p_{iz} \right] \times n_{ix}^M \\
&\quad + \left[ (-c\alpha s\gamma + s\alpha s\beta c\gamma) p_{iy} + (-c\alpha c\gamma - s\alpha s\beta s\gamma) p_{iz} \right] \times n_{iy}^M \\
&\quad + \left[ c\beta c\gamma p_{iy} - c\beta s\gamma p_{iz} \right] \times n_{iz}^M, \\
&\boldsymbol{J}_{riy}^p = \left[ -c\alpha s\beta p_{ix} + (c\alpha c\beta s\gamma) p_{iy} + (c\alpha c\beta c\gamma) p_{iz} \right] \times n_{ix}^M \\
&\quad + \left[ -s\alpha s\beta p_{ix} + (s\alpha c\beta s\gamma) p_{iy} + (s\alpha c\beta c\gamma) p_{iz} \right] \times n_{iy}^M \\
&\quad + \left[ -c\beta p_{ix} - s\beta s\gamma p_{iy} - s\beta c\gamma p_{iz} \right] \times n_{iz}^M, \\
&\boldsymbol{J}_{riz}^{p} = [ -s \alpha c \beta p_{ix} + (-c \alpha c \gamma - s \alpha s \beta s \gamma) p_{iy} \\
&\quad + (c \alpha s \gamma - s \alpha s \beta c \gamma) p_{iz} ] \times n_{ix}^{M} \\
&\quad + [ c \alpha c \beta p_{ix} + (-s \alpha c \gamma + c \alpha s \beta s \gamma) p_{iy} \\
&\quad + (s \alpha s \gamma + c \alpha s \beta c \gamma) p_{iz} ] \times n_{iy}^{M}.
\end{aligned}
\end{equation*}

Subsequent calculations for \(\boldsymbol{J}_{ti}^p\) and \(\boldsymbol{J}_{ri}^p\) in this paper will use \eqref{equa24} and \eqref{equa26}, respectively.

\subsubsection{Localizability Contribution Vector}
The Jacobian in the rotation direction is influenced by the magnitude of the coordinates of point \(\boldsymbol{p}\), which often leads to larger values.  To facilitate the subsequent setting of uniform thresholds for translation and rotation directions, the scales of localizability in these two directions are unified. Here, we adopt the matrix normalization method from \cite{Tuna2024_xicp} to normalize the results in the rotation direction. If \(\left\| \boldsymbol{J}_{ri}^{e} \right\|\) in \eqref{equa23} or \(\left\| \boldsymbol{J}_{ri}^{p} \right\|\) in \eqref{equa26} is greater than 1, the operation is performed as follows:

\begin{equation*} 
\label{equa_normalizatione}
\boldsymbol{J}_{ri}^{e} = 
\begin{cases}
\frac{\boldsymbol{J}_{ri}^{e}}{\left\|\boldsymbol{J}_{ri}^{e}\right\|}, & \text{if} \ \left\|\boldsymbol{J}_{ri}^{e}\right\| > 1 \\
\boldsymbol{J}_{ri}^{e}, & \text{otherwise}
\end{cases}
\end{equation*}

\begin{equation*}
\label{equa_normalizationp}
\boldsymbol{J}_{ri}^{p} = 
\begin{cases}
\frac{\boldsymbol{J}_{ri}^{p}}{\left\|\boldsymbol{J}_{ri}^{p}\right\|}, & \text{if} \ \left\|\boldsymbol{J}_{ri}^{p}\right\| > 1 \\
\boldsymbol{J}_{ri}^{p}, & \text{otherwise}
\end{cases}
\end{equation*}

Finally, based on \eqref{equa18} and \eqref{equa20}, combined with \eqref{equa15}, \eqref{equa16}, \eqref{equa21}, and \eqref{equa23}, the localizability contribution vector of the edge point \(\boldsymbol{p}_i\) to the line correspondence constraint can be obtained as follows:

\begin{equation} 
\label{equa27}
\boldsymbol{F}_{ri}^{e} = \begin{pmatrix} (\boldsymbol{J}_{ri}^{e}  \boldsymbol{v}_{r1})^2 \\ (\boldsymbol{J}_{ri}^{e}  \boldsymbol{v}_{r2})^2 \\ (\boldsymbol{J}_{ri}^{e}  \boldsymbol{v}_{r3})^2 \end{pmatrix} \in \mathbb{R}^{3 \times 1}
\end{equation}

\begin{equation} 
\label{equa28}
\boldsymbol{F}_{ti}^{e} = \begin{pmatrix} (\boldsymbol{J}_{ti}^{e}  \boldsymbol{v}_{t1})^2 \\ (\boldsymbol{J}_{ti}^{e}  \boldsymbol{v}_{t2})^2 \\ (\boldsymbol{J}_{ti}^{e}  \boldsymbol{v}_{t3})^2 \end{pmatrix} \in \mathbb{R}^{3 \times 1}
\end{equation}

Similarly, based on \eqref{equa18} and \eqref{equa20}, combined with \eqref{equa15}, \eqref{equa16}, \eqref{equa24}, and \eqref{equa26}, the localizability contribution vector of the planar point \(\boldsymbol{p}_j\) to the plane correspondence constraint can be obtained as follows:

\begin{equation} 
\label{equa29}
\boldsymbol{F}_{ri}^{p} = \begin{pmatrix} (\boldsymbol{J}_{ri}^{p}  \boldsymbol{v}_{r1})^2 \\ (\boldsymbol{J}_{ri}^{p}  \boldsymbol{v}_{r2})^2 \\ (\boldsymbol{J}_{ri}^{p}  \boldsymbol{v}_{r3})^2 \end{pmatrix} \in \mathbb{R}^{3 \times 1}
\end{equation}

\begin{equation} 
\label{equa30}
\boldsymbol{F}_{ti}^{p} = \begin{pmatrix} (\boldsymbol{J}_{ti}^{p}  \boldsymbol{v}_{t1})^2 \\ (\boldsymbol{J}_{ti}^{p}  \boldsymbol{v}_{t2})^2 \\ (\boldsymbol{J}_{ti}^{p}  \boldsymbol{v}_{t3})^2 \end{pmatrix} \in \mathbb{R}^{3 \times 1}
\end{equation}

All the localizability contribution vectors related to edge points and planar points are stacked together to form the following information matrix:

\begin{equation*} 
\label{equa31}
\boldsymbol{F}_r = \left[ \boldsymbol{F}_{r1}^{e} \cdots \boldsymbol{F}_{rN_1}^{e} \; \boldsymbol{F}_{r1}^{p} \cdots \boldsymbol{F}_{rN_2}^{p} \right]^T
\end{equation*}

\begin{equation*} 
\label{equa32}
\boldsymbol{F}_t = \left[ \boldsymbol{F}_{t1}^{e} \cdots \boldsymbol{F}_{tN_1}^{e} \; \boldsymbol{F}_{t1}^{p} \cdots \boldsymbol{F}_{tN_2}^{p} \right]^T
\end{equation*}

\subsection{Determining the Categories of Localizability}
\label{section:detection_module_C}
All the localizability contribution vectors related to edge points and planar points have been obtained. Next, these vectors will be used to determine the localizability categories for each direction of the 6-DOF pose. 

First, in order to reduce the influence of noise, low localizability contributions will be filtered out. At the same time, higher localizability contributions will be selected for subsequent constraint calculations. Furthermore, the localizability in each direction of the 6-DOF pose will be summed and compared with thresholds to determine the corresponding localizability category for each direction.

\subsubsection{Filtering Out Noise and Picking Out High Localizability Contributions}
Define \(\boldsymbol{F} = \left[ \boldsymbol{F}_r, \boldsymbol{F}_t \right] \in \mathbb{R}^{N \times 6}\) as the localizability matrix. In order to filter out low localizability contributions and reduce the interference of noise, we set a low contribution threshold \(h_f\). For each localizability component \(F(i,\,j)\) in \(\boldsymbol{F}\) (i.e. the localizability contribution of the i-th correspondence constraint in the direction \(\boldsymbol{v}_j\)), if it is lower than \(h_f\), it is set to 0 and considered as noise, as shown in \eqref{equa33}. Then, we further distinguish between moderate localizability contributions \(F_f\) and high localizability contributions \(F_u\).

\begin{equation} 
\label{equa33}
F_f(i,j) = 
\begin{cases} 
F(i,j), & \text{if } F(i,j) \geq h_f \\
0, & \text{otherwise}
\end{cases}
\end{equation}

This paper empirically suggests \(h_f=0.03\). The value of \(h_f\) can be adjusted based on the noise characteristics of the LiDAR. The greater the noise, the larger the value. Additionally, non-zero localizability contributions in \(F_f\) can be considered as moderate localizability contributions. The sum of all moderate localizability contributions in the direction of \(\boldsymbol{v}_j\) is calculated as follows:

\begin{equation} 
\label{equa34}
L_f(j) = \sum_{i=1}^{N_1+N_2} F_f(i, j)
\end{equation}

\(L_f (j)\) is the sum of the localizability contributions of all useful correspondence constraints in the  direction of \(\boldsymbol{v}_j\). The larger its value, the stronger the overall localizability contribution in the direction, indicating less degeneracy.

Next, we set a high contribution threshold \(h_u\). Localizability contributions in \(F_f(i, j)\) greater than \(h_u\) are considered high localizability contributions, indicating that the corresponding correspondence constraint has a strong influence in the direction of \(\boldsymbol{v}_j\), as follows:

\begin{equation} 
\label{equa35}
F_u(i,j) = 
\begin{cases} 
F(i,j), & \text{if } F(i,j) \geq h_u \\
0, & \text{otherwise}
\end{cases}
\end{equation}

This paper empirically suggests \(h_u=0.4998\).

The sum of all high localizability contributions in the direction of \(\boldsymbol{v}_j\) is calculated as follows:

\begin{equation} 
\label{equa36}
L_u(j) = \sum_{i=1}^{N_1+N_2} F_u(i, j)
\end{equation}

\(L_u(j)\) is the sum of the localizability contributions of all high localizability points in the direction of \(\boldsymbol{v}_j\). The larger its value, it indicates that more high-quality constraints can be provided in that direction.

\begin{figure*}[t]
\centering
\includegraphics[width=1.0\linewidth]{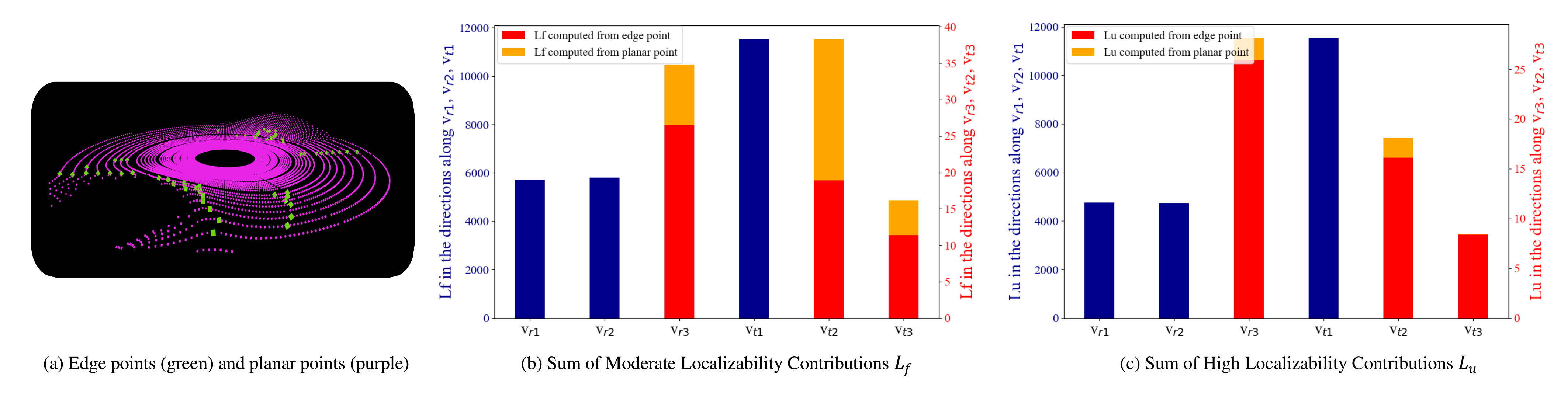}
\caption{(a) Edge points (green) and planar points (purple) extracted by the front end of LVI-SAM\cite{Shan2021_lvisam}. These points are from a LiDAR scan obtained in an open area with ground undulations. (b)-(c) Two exemplary double Y-axis histograms presenting \(L_f\) and   \(L_u\) along each eigenvector direction of the 6-DOF pose. The localizability contributions are computed from the valid points shown in (a). It indicates smaller values of \(L_f(j)\) and   \(L_u(j)\) in the \(\boldsymbol{v}_{r3}\), \(\boldsymbol{v}_{t2}\) and \(\boldsymbol{v}_{t3}\) directions. The red region represents the localizability contribution from edge point-to-line correspondences, while the orange region represents the contribution from planar point-to-plane correspondences.}
\label{fig:histogram}
\end{figure*}

An example of the calculation results for \(L_f(j)\) and \(L_u(j)\) is provided in Fig.~\ref{fig:histogram}. They are computed from the localizability contributions of a LiDAR scan performed in an open area with slightly undulating terrain.  Fig.~\ref{fig:histogram}(a) shows the edge points and planar points extracted from this LiDAR scan. Figs.~\ref{fig:histogram}(b) and (c) present the corresponding calculation results. In the directions along \(\boldsymbol{v}_{r1}\), \(\boldsymbol{v}_{r2}\) and \(\boldsymbol{v}_{t1}\), the values of \(L_f(j)\) and   \(L_u(j)\) are large, indicating well-constrained conditions. In the directions along  \(\boldsymbol{v}_{r3}\), \(\boldsymbol{v}_{t2}\) and \(\boldsymbol{v}_{t3}\), the values of \(L_f(j)\) and   \(L_u(j)\) are small, indicating ill-conditioned constraints. In ill-conditioned directions, the calculation of \(L_u(j)\) which represents the sum of high localizability contributions, is dominated by the contributions from edge point-to-line correspondences (red region). Therefore, it is essential to utilize the localizability information from edge point-to-line correspondences.

\subsubsection{Categorization}
The localizability contributions of all correspondence constraints related to edge points and planar points along the directions of the eigenvectors in the 6-DOF pose are summed to determine the localizability category for each corresponding direction. \(L_f(j)\) represents the overall localizability contribution in the direction of \(\boldsymbol{v}_j\). \(L_u(j)\) represents the localizability contribution of high-quality constraints in the direction of \(\boldsymbol{v}_j\). Based on their characteristics, thresholds \(T_1\), \(T_2\), \(T_3\) and \(T_4\) are set to determine the localizability category of the 6-DOF pose along the eigenvectors directions as follows:

in the direction along \(\boldsymbol{v}_j\), 

\begin{equation} 
\label{equa37}
\text{Category} = 
\begin{cases} 
\text{Full}, & \text{if } L_f(j) \geq T_1 \text{ or } L_u(j) \geq T_2 \\
\text{Partial}, & \text{if } L_f(j) \geq T_3 \text{ and } L_u(j) \geq T_4 \\
\text{None}, & \text{otherwise}
\end{cases}
\end{equation}

That is, when \(L_f(j) \geq T_1 \text{ or } L_u(j) \geq T_2\), the localizability category of the direction along \(\boldsymbol{v}_j\) is considered fully localizable, i.e. the category is ``Full". When \(L_f(j) \geq T_3 \text{ and } L_u(j) \geq T_4\), the localizability category of the direction along \(\boldsymbol{v}_j\) is considered partially localizable, i.e. the category is ``Partial". In this direction, both \(L_f(j)\) and  \(L_u(j)\) must meet their respective thresholds. This means that there must be a sufficient number of points with high localizability contributions in the direction along \(\boldsymbol{v}_j\). Otherwise, the localizability category of the direction of \(\boldsymbol{v}_j\) is considered non-localizable, i.e. the category is ``None". In this study, based on experimental validation, for subsequent experiments, \(T_1\) is set to 50, \(T_2\) is set to 30, \(T_3\) is set to 15 and \(T_4\) is set to 9.

\begin{figure*}[t]
\centering
\includegraphics[width=1.0\linewidth]{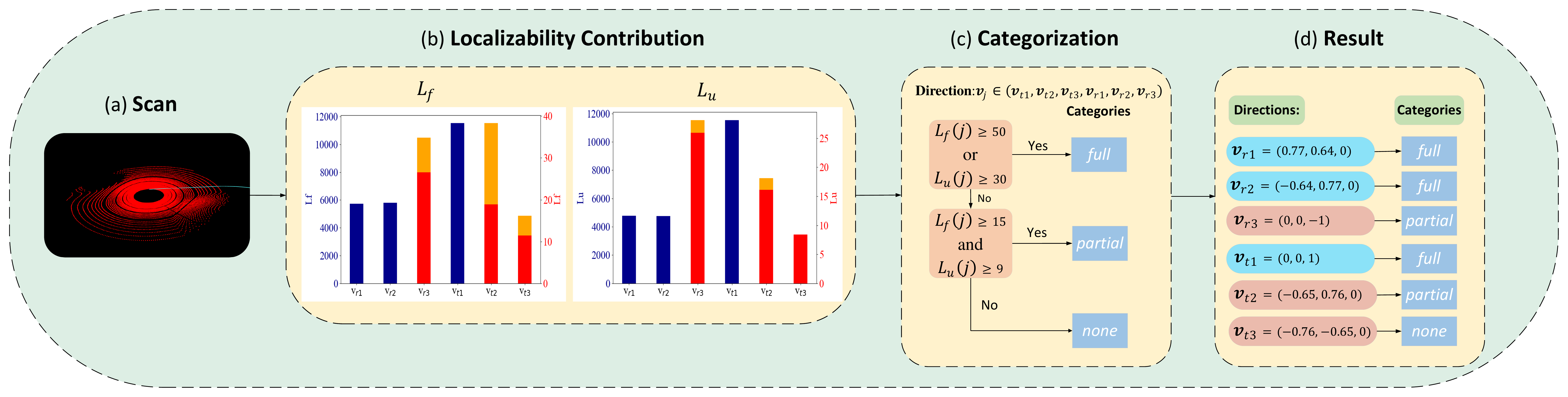}
\caption{An example of the Localizability Detection Module. (a) Scan: A LiDAR scan obtained in an open area with slightly undulating terrain. The same point cloud as shown in Fig.~\ref{fig:histogram} (a) is visualized here from a different viewpoint. (b) Localizability Contribution: Two double Y-axis histograms identical to those in Fig.~\ref{fig:histogram}. They are calculated based on the localizability contribution of the scan in (a). (c) Categorization: The localizability category along each eigenvector direction of the 6-DOF pose is assigned through conditional judgment. The threshold is set according to Section~\ref{section:detection_module_C}. (d)Result: The results of the localizability categories along the eigenvector directions  are shown.}
\label{fig:locDetect_exam}
\end{figure*}

The LocDetect-Module uses a scan from LiDAR to assign the localizability category of the 6-DOF pose along each eigenvector direction. An example is shown in Fig.~\ref{fig:locDetect_exam}, which illustrates the following process: In an open area with undulating terrain, the current scan from LiDAR is used as input. By calculating the localizability contributions of all edge point-to-line and planar point-to-plane correspondences, the values of \(L_f(j)\) and \(L_u(j)\) in the direction along  \(\boldsymbol{v}_j\) are obtained. These values are presented in the same double y-axis histograms as in Fig.~\ref{fig:histogram}, shown in Fig.~\ref{fig:locDetect_exam}(b). Subsequently, the values of \(L_f(j)\) and \(L_u(j)\) for each direction are conditionally evaluated and compared with predefined thresholds to determine the localizability category for that direction. The decision-making process is illustrated in Fig.~\ref{fig:locDetect_exam}(c). The categorization results for all directions are shown in Fig.~\ref{fig:locDetect_exam}(d) as follows: 

The categories for directions along \(\boldsymbol{v}_{r1}\), \(\boldsymbol{v}_{r2}\) and \(\boldsymbol{v}_{t1}\) are fully localizable, indicating good geometric constraints. the categories for directions along \(\boldsymbol{v}_{r3}\), \(\boldsymbol{v}_{t2}\) are partially localizable, suggesting that although these directions exhibit degeneracy, some geometric information may still be usable. The category for directions along \(\boldsymbol{v}_{t3}\) is non-localizable, indicating severe degeneracy in this direction with minimal usable geometric information.

\section{Optimization Module}
\label{section:optimization_module}
The Optimization Module utilizes the localizability information provided by the LocDetect-Module. Based on the localizability categories of the 6-DOF pose along the directions of the eigenvectors, additional soft and hard constraints are incorporated into the optimization equation. Specifically, in partially localizable directions, soft constraints are added using constraint values computed from higher localizability contributions. In non-localizable directions, hard constraints are applied to prevent pose updates. In fully localizable directions, pose updates remain unrestricted.

Note that to improve time efficiency in the subsequent experiments, the localizability category determination and additional constraint calculations are performed only during the first iteration of each ICP execution. For all subsequent iterations before convergence, the categories and additional constraint information computed in the first iteration are used directly in the optimization. Because in most SLAM algorithms, ICP-based pose estimation is typically performed at a relatively high frequency (e.g. 10 Hz in LVI-SAM), this configuration has little impact on accuracy.

\subsection{Optimization Strategy}
\label{section:optimization_module_A}
Binary degeneracy handling methods may omit usable constraint information, and the use of standalone hard constraints can sometimes induce fluctuations in the trajectory along partially localizable directions. To address the above issue, we use an optimization method that combines soft and hard constraints in the Optimization-Module, utilizing the localizability information from the LocDetect-Module. Each time the ICP algorithm starts, the process begins with the initial pose estimate \(\boldsymbol{x}_0\), which can be obtained through various methods in SLAM, such as previous odometry. Through iteration, the goal is to find the optimal pose update \(\Delta\boldsymbol{x}\), which is added to \(\boldsymbol{x}_0\) to obtain the optimal pose solution \(\boldsymbol{x}^*\).

Specifically, the optimization strategy is as follows:

In the case of the localizability category in the direction along \(\boldsymbol{v}_{j}\), if it is:

\begin{enumerate}
    \item Fully Localizable (Full): No additional constraints are added in this direction. The pose is updated without any extra constraints  in this direction.
    \item Partially Localizable (Partial): First, obtain the edge point-to-line correspondences and planar point-to-plane correspondences with moderate localizability contributions in the direction of \(\boldsymbol{v}_{j}\) from the LocDetect-Module (i.e., satisfying \eqref{equa33}). Then, use them for a small local ICP optimization, which combines point-to-line and point-to-plane cost functions, to obtain an additional constraint value \(\Delta \boldsymbol{r}_0 \in \mathbb{R}^{3 \times 1}\) or \(\Delta \boldsymbol{t}_0 \in \mathbb{R}^{3 \times 1}\) for the pose update in the direction of \(\boldsymbol{v}_{j}\). Then, a soft constraint is added with weights. The specific process is described in Section~\ref{section:optimization_module_B}.
    \item Non-localizable (None): The information available in the  direction of \(\boldsymbol{v}_{j}\) is scarce, so a hard constraint is added to make the pose update value zero. The specific process is described in Section~\ref{section:optimization_module_C}.
\end{enumerate}

\subsection{Adding Soft Constraints in Partially Localizable Directions}
\label{section:optimization_module_B}
According to the LocDetect-Module, if \(L_f(j) \geq T_3 \text{ and } L_u(j) \geq T_4\) hold in the direction of \(\boldsymbol{v}_{j}\), it is classified as a partially localizable category. In this direction, geometric constraints are not as abundant as in fully localizable directions, but there is still usable information. First, pick out all edge point-to-line and planar point-to-plane correspondences in this direction used for calculating \(L_f(j)\) in \eqref{equa34}. They are correspondences with moderate localizability contributions. Then, the constraint value is calculated. We refer to the method used in \cite{Tuna2024_xicp} and \cite{Shan2018_legoloam} for constraint calculation, which optimizes the pose in that direction by using correspondences that contribute more in that direction. A small ICP optimization is performed using these correspondences, which combines point-to-line and point-to-plane loss functions. If \(\boldsymbol{v}_{j}\) is an eigenvector in the rotational direction, the small ICP optimizes only the 3-DOF rotational state variables \(\boldsymbol{r}\). If \(\boldsymbol{v}_{j}\) is an eigenvector in the translational direction, the small ICP optimizes only the 3-DOF translational state variables \(\boldsymbol{t}\). The pose update estimate \(\Delta \boldsymbol{r}_0\) or \(\Delta \boldsymbol{t}_0\) obtained from this optimization is projected onto the direction of \(\boldsymbol{v}_{j}\) and treated as a constraint value for the final pose update in the direction of \(\boldsymbol{v}_{j}\). A soft constraint is then added to the pose update optimization equation \eqref{equa2} to guide the pose update in the direction of \(\boldsymbol{v}_{j}\) toward the constraint value. The algorithm process is as follows:

\begin{enumerate}
    \item According to the category information from the LocDetect-Module, the category in the direction of \(\boldsymbol{v}_{j}\) is partially localizable.
    \item Pick out all the \(m\) edge point-to-line correspondences and \(n\) planar point-to-plane correspondences in this direction, which are used to calculate \(L_f(j)\) in \eqref{equa34}. They are correspondences with moderate localizability contributions.
    \item Using the correspondences selected in step 2, solve the following small ICP problem using the Gauss-Newton method.

    When \(\boldsymbol{v}_{j}\) corresponds to a rotational direction, solve the following optimization problem to update only the 3-DOF rotation \(\boldsymbol{r}\).

\begin{equation} 
\label{equa38}
\min_{\Delta \boldsymbol{r}} \sum_{i=1}^{m} \left\| f_{ei}(\boldsymbol{r} + \Delta \boldsymbol{r}) \right\|^2 + \sum_{j=1}^{n} \left\| f_{pj}(\boldsymbol{r} + \Delta \boldsymbol{r}) \right\|^2
\end{equation}

Starting from the initial estimate, iterate to solve for the update \(\Delta \boldsymbol{r}\) and find the optimal update value \(\Delta \boldsymbol{r}_0\) that minimizes the residual, which will be used as the constraint value. 

When \(\boldsymbol{v}_{j}\) corresponds to a translational direction, solve the following optimization problem to update only the 3-DOF translation \(\boldsymbol{t}\).

\begin{equation} 
\label{equa39}
\min_{\Delta \boldsymbol{t}} \sum_{i=1}^{m} \left\| f_{ei}(\boldsymbol{t} + \Delta \boldsymbol{t}) \right\|^2 + \sum_{j=1}^{n} \left\| f_{pj}(\boldsymbol{t} + \Delta \boldsymbol{t}) \right\|^2
\end{equation}

Starting from the initial estimate, iterate to solve for the update \(\Delta \boldsymbol{t}\) and find the optimal update value \(\Delta \boldsymbol{t}_0\) that minimizes the residual, which will be used as the constraint value. 

Then, \(\Delta \boldsymbol{r}_0\) or \(\Delta \boldsymbol{t}_0\), along with \(\boldsymbol{v}_{j}\), are each extended into a six-dimensional vector.

If  \(\boldsymbol{v}_{j}\) corresponds to a rotational direction, as follows:

\begin{equation} 
\label{equa40}
\boldsymbol{v}_j' = \begin{pmatrix} \boldsymbol{v}_j \\ \boldsymbol{0}_{3 \times 1} \end{pmatrix}, \quad \Delta \boldsymbol{x}_j' = \begin{pmatrix} \Delta \boldsymbol{r}_0 \\ \boldsymbol{0}_{3 \times 1} \end{pmatrix}.
\end{equation}

If  \(\boldsymbol{v}_{j}\) corresponds to a translational direction, as follows:

\begin{equation} 
\label{equa41}
\boldsymbol{v}_j' = \begin{pmatrix} \boldsymbol{0}_{3 \times 1} \\ \boldsymbol{v}_j \end{pmatrix}, \quad \Delta \boldsymbol{x}_j' = \begin{pmatrix} \boldsymbol{0}_{3 \times 1} \\ \Delta t_0 \end{pmatrix}.
\end{equation}

\item A soft constraint is added to the final optimization equation, encouraging the final pose update in the direction of \(\boldsymbol{v}_{j}\) to approach the projection of the constraint value \(\Delta \boldsymbol{r}_0\) or \(\Delta \boldsymbol{t}_0\) in the direction of \(\boldsymbol{v}_{j}\). A squared loss function is used, as follows:

\begin{equation} 
\label{equa42}
\mu_j \left( \boldsymbol{v}_j'^T \Delta \boldsymbol{x} - \boldsymbol{v}_j'^T \Delta \boldsymbol{x}_j' \right)^2
\end{equation}
where \(\mu_j = 
\begin{cases} 
2, & \text{if } L_u(j) < T_5 \\
5, & \text{if } L_u(j) \geq T_5 .
\end{cases}\)
    
\end{enumerate}

\(\mu_j\) is the set weight. By adjusting the value of \(\mu_j\), the extent to which the final pose update's projection in the \(\boldsymbol{v}_{j}\) direction tends toward the projection of the constraint values \(\Delta \boldsymbol{r}_0\) or \(\Delta \boldsymbol{t}_0\) in the \(\boldsymbol{v}_{j}\) direction can be controlled. In this paper, \(T_5\) is set to 15 for all subsequent experiments. When \(L_u(j)\) in the direction of \(\boldsymbol{v}_{j}\) is greater than \(T_5\), it indicates that more correspondences with high localizability contributions are used to compute \(\Delta \boldsymbol{r}_0\) or \(\Delta \boldsymbol{t}_0\), making the calculated values more reliable, and thus the weight \(\mu_j\) is increased. If the quality and quantity of the correspondences used in \eqref{equa38} or \eqref{equa39} are high, the weight \(\mu_j\) can be appropriately increased. However, if \(\mu_j\) is too large, it may cause fluctuations and jaggedness in the optimized trajectory, leading to instability. The setting of \(\mu_j\) should ensure that the pose estimate in the direction of \(\boldsymbol{v}_{j}\) tends toward the constraint value, while also maintaining the smoothness and stability of the trajectory. Nonetheless, the use of soft constraints in partially localizable directions may introduce some uncertainty in the estimation results, and their impact will be further investigated in future work.

Soft constraint terms in \eqref{equa42} are added to all directions \(\boldsymbol{v}_{j_i}'\) categorized as partially locatable. Note that \(\boldsymbol{v}_{j_i}'\) here refers to the direction vector of the i-th partially locatable direction. If there are \(k_p\) partially locatable directions (\(k_p \leq 6\)), the following soft constraints are added to the final optimization equation:

\begin{equation} 
\label{equa43}
\sum_{i=1}^{k_p} \mu_{j_i} \left( \boldsymbol{v}_{j_i}'^T \Delta \boldsymbol{x} - \boldsymbol{v}_{j_i}'^T \Delta \boldsymbol{x}_{j_i}' \right)^2
\end{equation}
where \(\Delta \boldsymbol{x}_{j_i}'\) here refers to the constraint \(\Delta \boldsymbol{x}_{j}'\) for the i-th partially locatable direction.

\subsection{Adding Hard Constraints in Non-Locatable Directions}
\label{section:optimization_module_C}
For the direction along \(\boldsymbol{v}_{j_i}'\), categorized as non-locatable, a hard constraint is added to ensure that the projection of the final pose update along \(\boldsymbol{v}_{j_i}'\) is zero, as follows:

\begin{equation} 
\label{equa44}
\boldsymbol{v}_j'^T \Delta \boldsymbol{x} = 0
\end{equation}

Note that \(\boldsymbol{v}_j' \in \mathbb{R}^{6 \times 1}\) is extended from \(\boldsymbol{v}_j \in \mathbb{R}^{3 \times 1}\), with the extension method similar to \eqref{equa40} or \eqref{equa41}.

The final optimization equation incorporates the hard constraint in \eqref{equa44} for all direction \(\boldsymbol{v}_{j_i}'\) categorized as non-localizable. Note that \(\boldsymbol{v}_{j_i}'\) here denotes the direction vector of the i-th non-localizable direction. If there are \(k_n\) directions categorized as non-localizable (\(k_n\leq 6\)), the following hard constraints are added to the final optimization equation:

\begin{equation} 
\label{equa45}
\boldsymbol{D}_{k_n \times 6} \Delta \boldsymbol{x} = 0
\end{equation}
where \(\boldsymbol{D}_{k_n \times 6} = \begin{pmatrix} \boldsymbol{v}_{j_1}' & \cdots & \boldsymbol{v}_{j_{k_n}}' \end{pmatrix}^T\).

\subsection{Optimization Equation with Soft and Hard Constraints}
\label{section:optimization_module_D}
\subsubsection{Optimization Equation}
Formulate the optimization equation. Then, Starting from the initial estimate, the pose update value \(\Delta\boldsymbol{x}\) is iteratively solved using the Gauss-Newton method to minimize the residual. The optimal update value  \(\Delta\boldsymbol{x}\) is then added to the initial estimate to obtain the optimal pose solution  \(\boldsymbol{x}^*\).

By combining \eqref{equa2}, \eqref{equa43}, and \eqref{equa45}, the final optimization equation with soft and hard constraints is obtained as follows:

\begin{equation} 
\label{equa46}
\begin{aligned}
& \min_{\Delta \boldsymbol{x} \in \mathbb{R}^6} \sum_{i=1}^{N} \| f_i(\boldsymbol{x} + \Delta \boldsymbol{x}) \|^2 + \sum_{i=1}^{k_p} \mu_{j_i} \left( \boldsymbol{v}_{j_i}'^T \Delta \boldsymbol{x} - \boldsymbol{v}_{j_i}'^T \Delta \boldsymbol{x}_{j_i}' \right)^2 \\
& \quad \quad \quad \quad \quad \quad \quad \quad \text{s.t.} \quad \boldsymbol{D} \Delta \boldsymbol{x} = 0.
\end{aligned}
\end{equation}

Here, let \(\mu_j = 
\begin{cases} 
2, & \text{if } L_u(j) < T_5 \\
5, & \text{if } L_u(j) \geq T_5 
\end{cases}\), with \(T_5=15\).

\begin{algorithm}[t]
\caption{Registration Process of LP-ICP} \label{alg:algorithm1}
\begin{algorithmic}[1]
\STATE \textbf{Input:} initial pose estimate \( \boldsymbol{x}_0 \), Lidar scan point cloud \( P^L \) (edge and planar points extracted), Map point cloud \( Q \) from the mapping module, Number of edge points \( N_1 \), number of planar points \( N_2 \)
\STATE \textbf{Output:} Optimal pose estimate \( \boldsymbol{x}^* \)
\STATE \( \boldsymbol{x}^* \gets \boldsymbol{x}_0 \);
\WHILE{nonlinear iterations}
    \STATE \( P^M \gets \boldsymbol{T}(\boldsymbol{x}^*) \cdot P^L \);
    \FOR{each point in \( P^M \)}
        \STATE search for matching points in \( Q \);
        \STATE Calculate  \( f_i(\boldsymbol{x}) \) and \( \boldsymbol{J}_i(\boldsymbol{x}) \);
    \ENDFOR
    \STATE Construct \( \boldsymbol{H}\),  \(\boldsymbol{H}_r \), \( \boldsymbol{H}_t \), \( \boldsymbol{V}_r \) and \(\boldsymbol{V}_t \);
    \IF{first iteration}
        \FOR{\(i=1\) to \( N_1+N_2\)}
            \IF{ \(i<= N_1\)}
                 \STATE Compute \( \boldsymbol{F}_{ri}^e \) and \( \boldsymbol{F}_{ti}^e \) based on \eqref{equa27} and \eqref{equa28};
            \ELSE
                \STATE Compute \( \boldsymbol{F}_{ri}^p \) and \( \boldsymbol{F}_{ti}^p \) based on \eqref{equa29} and \eqref{equa30};
            \ENDIF
            \STATE Update \( \boldsymbol{F}_t \) and \( \boldsymbol{F}_r \);
            \STATE Compute \( F_f(i,j) \) and \( F_u(i,j) \) using \eqref{equa33} and \eqref{equa35};\STATE \( L_f(j) \gets L_f(j) + F_f(i,j) \);
            \STATE \( L_u(j) \gets L_u(j) + F_u(i,j) \);
        \ENDFOR
        \FOR{each \( \boldsymbol{v}_j \in \{\boldsymbol{v}_{r1}, \boldsymbol{v}_{r2}, \boldsymbol{v}_{r3}, \boldsymbol{v}_{t1}, \boldsymbol{v}_{t2}, \boldsymbol{v}_{t3}\} \)}
            \STATE Determine localizability category based on \eqref{equa37};
            \IF{Category == partial}
                \STATE Perform \eqref{equa38} or \eqref{equa39} to obtain \( \Delta \boldsymbol{x}_{(j_i)}' \);
                \STATE Compute Soft constraints based on \eqref{equa43};
            \ELSIF{Category == none}
                \STATE Add \( \boldsymbol{v}_{(j_i)}' \) to \( D \);
            \ENDIF
        \ENDFOR
    \ENDIF
    \STATE Construct  \( \boldsymbol{H}'\) and  \( \boldsymbol{b}\);
    \STATE Solve for \( \Delta \boldsymbol{x} \) based on \eqref{equa47};
    \STATE \( \boldsymbol{x}^* \gets \boldsymbol{x}^* + \Delta \boldsymbol{x} \);
\ENDWHILE
\RETURN \( \boldsymbol{x}^* \)
\end{algorithmic}
\end{algorithm}

\subsubsection{Solve}
Using Lagrange multipliers methods, the constrained optimization problem \eqref{equa46} is transformed into an unconstrained optimization problem. The pose update value can then be obtained by solving it using the Gauss-Newton method, as follows:

\begin{equation} 
\label{equa47}
\begin{aligned}
&\left(\begin{array}{cc}
\boldsymbol{H}' & \boldsymbol{D}^T \\
\boldsymbol{D} & \boldsymbol{0}
\end{array}\right)
\left(\begin{array}{c}
\Delta \boldsymbol{x} \\
\boldsymbol{\lambda}
\end{array}\right)
=
\left(\begin{array}{c}
\boldsymbol{b} \\
\boldsymbol{0}
\end{array}\right) \\
&\text{where} \quad \boldsymbol{\lambda} \in \mathbb{R}^{k_n \times 1}, \\
&\boldsymbol{H}' = 2 \sum_{i=1}^{N} \boldsymbol{J}_i^T(\boldsymbol{x}) \boldsymbol{J}_i(\boldsymbol{x}) + 2 \sum_{i=1}^{k_p} \mu_{j_i} \boldsymbol{v}_{j_i}' \boldsymbol{v}_{j_i}'^T ,\\
&\boldsymbol{b} = -2 \sum_{i=1}^{N} \boldsymbol{J}_i^T(\boldsymbol{x}) f_i(\boldsymbol{x}) + 2 \sum_{i=1}^{k_p} \mu_{j_i} \boldsymbol{v}_{j_i}' \boldsymbol{v}_{j_i}'^T \Delta \boldsymbol{x}_{j_i}'.
\end{aligned}
\end{equation}

Here, \(\boldsymbol{\lambda} \in \mathbb{R}^{k_n \times 1}\) are Lagrangian multipliers. \(\boldsymbol{J}_i(\boldsymbol{x})\in \mathbb{R}^{1 \times 6}\) is the Jacobian of the residual \(\boldsymbol{f}_i(\boldsymbol{x})\) w.r.t. \(\boldsymbol{x}\).

Eq. \eqref{equa47} can be solved using QR decomposition to obtain the pose update \(\Delta\boldsymbol{x}\) which is then added to the initial estimate. Through multiple iterations, the optimal pose solution \(\boldsymbol{x}^*\) is obtained. The proposed algorithm is summarized in Algorithm~\ref{alg:algorithm1}.

\section{Experiment}
\label{section:experiment}
In this section, the experimental setup is introduced in Section~\ref{section:experiment_A}. The proposed method is then evaluated through experiments on a simulated planetary-like environment dataset (Section~\ref{section:experiment_B}), a real-world underground tunnel dataset (Section~\ref{section:experiment_C}) and the Long Corridor sequence from the SubT-MRS dataset (Section~\ref{section:experiment_subt}), with comparisons to state-of-the-art methods. Finally, the runtime performance of the algorithm is evaluated (Section~\ref{section:experiment_runtime}).

\subsection{Experimental Setup}
\label{section:experiment_A}
\subsubsection{Implementation Details}
The proposed framework, LP-ICP, is integrated into the point cloud registration framework of LVI-SAM \cite{ Shan2021_lvisam}. LVI-SAM combines LIO and VIO sub-modules, with the LIO sub-module serving as the main component, while the VIO sub-module provides initial estimates for the LIO sub-module. The LIO module performs scan-to-map registration using edge point-to-line and planar point-to-plane correspondences. The tested algorithms are run five times on each sequence, and the best result is reported for evaluation. All experiments and evaluations in this paper are performed on a laptop equipped with an Intel i7-12700H CPU.

\subsubsection{Algorithmic Comparisons}
We integrate the current state-of-the-art methods Zhang et al. \cite{Zhang2016_zhang_etal} and X-ICP \cite{Tuna2024_xicp} into the point cloud registration framework of LVI-SAM for experiments and comparisons. In LVI-SAM, the degeneracy detection method used is based on Zhang et al. \cite{Zhang2016_zhang_etal}. To ensure that \cite{Zhang2016_zhang_etal} performs well for degeneracy detection and handling in all experiments, we empirically set the threshold of Zhang et al. \cite{Zhang2016_zhang_etal} to 50. We also conducted experiments to evaluate its performance with different threshold settings, as shown in Table~\ref{table:ZhangThr}. Finally, we choose a threshold of 50 for the subsequent experimental evaluations. Additionally, we integrated X-ICP \cite{Tuna2024_xicp} into the point cloud registration framework of LVI-SAM, based on the paper and its open-source implementation\footnote{\url{https://github.com/leggedrobotics/perfectlyconstrained}}. After parameter tuning, the thresholds were empirically set to $\kappa_1 = 90$, $\kappa_2 = 50$, and $\kappa_3 = 35$. In our experiments, the version integrated into LVI-SAM is referred to as Xn-ICP. In addition, we also evaluated the performance of several advanced SLAM systems in our experiments.

Note that in all experiments, considering the algorithm's time efficiency, for the three methods being compared, the localizability categories of the directions are only detected during the first iteration of point cloud registration. However, the categories and added additional constraints are used in every iteration during the optimization.

In this paper, the three methods integrated into the LVI-SAM framework for comparison are referred to as: LVI-SAM + Ours, LVI-SAM + Zhang et al. \cite{Zhang2016_zhang_etal}, and LVI-SAM + Xn-ICP \cite{Tuna2024_xicp}.

\begin{table}[t]
\centering
\caption{The RMSE of ATE for LVI-SAM + Zhang et al. \cite{Zhang2016_zhang_etal} across different eigenvalue thresholds in the a2\_traverse sequence experiment}
\label{table:ZhangThr}
\begin{threeparttable}
\renewcommand{\arraystretch}{1.2} 
\begin{tabular}{>{\centering\arraybackslash}p{2.5cm} >{\centering\arraybackslash}p{2.5cm}}
\hline
\rule[-1.5ex]{0pt}{4ex} 
          & a2\_traverse \\ \hline
Thr = 15  & 37.66        \\
Thr = 50  & \textbf{36.60}        \\
Thr = 100 & 195.57       \\
Thr = 500 & 231.79      \\ \hline
\end{tabular}
\begin{tablenotes}[para,flushleft]
        \item The units are in meters.
        \item The bold value indicates the best result.
     \end{tablenotes} 
\end{threeparttable} 
\end{table}

\begin{table*}[ht]
\centering
\caption{The RMSE of ATE and Ratio of Completion of the Compared Algorithms on our PLAM Dataset}
\label{table:exp1result}
\begin{threeparttable}
\renewcommand{\arraystretch}{1.2} 
\begin{tabular}{ccccccccccc}
\hline
\rule[-1.5ex]{0pt}{4ex} 
                       & a2\_odom                                                 & a3\_odom                                                 & a4\_loop                                                & a6\_loop                                                 & a7\_loop                                                & a1\_home                                                 & a2\_home                                                 & a4\_home                                                 & a3\_final                                                & a2\_traverse                                             \\ \hline
Length of trajectory   & 900                                                      & 805                                                      & 1038                                                    & 890                                                      & 1244                                                    & 864                                                      & 879                                                      & 1943                                                     & 745                                                      & 2231                                                     \\ \hline
ORB-SLAM3 \cite{Campos2021_orbslam3}             & \begin{tabular}[c]{@{}c@{}}119.73\\ (100\%)\end{tabular} & \begin{tabular}[c]{@{}c@{}}78.37\\ (71\%)\end{tabular}   & \begin{tabular}[c]{@{}c@{}}92.33\\ (40\%)\end{tabular}  & \begin{tabular}[c]{@{}c@{}}106.13\\ (100\%)\end{tabular} & \begin{tabular}[c]{@{}c@{}}12.71\\ (21\%)\end{tabular}  & \begin{tabular}[c]{@{}c@{}}54.77\\ (50\%)\end{tabular}   & \begin{tabular}[c]{@{}c@{}}110.57\\ (58\%)\end{tabular}  & \begin{tabular}[c]{@{}c@{}}99.70\\ (52\%)\end{tabular}   & \begin{tabular}[c]{@{}c@{}}61.12\\ (100\%)\end{tabular}  & \begin{tabular}[c]{@{}c@{}}32.85\\ (24\%)\end{tabular}   \\ \hline
VINS-Mono \cite{Qin2018_vinsmono}             & \begin{tabular}[c]{@{}c@{}}98.03\\ (100\%)\end{tabular}  & \begin{tabular}[c]{@{}c@{}}327.55\\ (100\%)\end{tabular} & \begin{tabular}[c]{@{}c@{}}80.50\\ (44\%)\end{tabular}  & \begin{tabular}[c]{@{}c@{}}142.02\\ (100\%)\end{tabular} & \begin{tabular}[c]{@{}c@{}}135.79\\ (56\%)\end{tabular} & \begin{tabular}[c]{@{}c@{}}332.41\\ (100\%)\end{tabular} & \begin{tabular}[c]{@{}c@{}}228.75\\ (100\%)\end{tabular} & \begin{tabular}[c]{@{}c@{}}247.04\\ (100\%)\end{tabular} & \begin{tabular}[c]{@{}c@{}}114.54\\ (100\%)\end{tabular} & \begin{tabular}[c]{@{}c@{}}202.66\\ (100\%)\end{tabular} \\ \hline
LIO-SAM \cite{Shan2020_liosam}               & \begin{tabular}[c]{@{}c@{}}9.22\\ (67\%)\end{tabular}    & \begin{tabular}[c]{@{}c@{}}31.72\\ (34\%)\end{tabular}   & \begin{tabular}[c]{@{}c@{}}1.65\\ (20\%)\end{tabular}   & \begin{tabular}[c]{@{}c@{}}0.61\\ (100\%)\end{tabular}   & \begin{tabular}[c]{@{}c@{}}6.51\\ (77\%)\end{tabular}   & \begin{tabular}[c]{@{}c@{}}23.68\\ (41\%)\end{tabular}   & \begin{tabular}[c]{@{}c@{}}15.62\\ (100\%)\end{tabular}  & \begin{tabular}[c]{@{}c@{}}29.41\\ (75\%)\end{tabular}   & \begin{tabular}[c]{@{}c@{}}1.10\\ (28\%))\end{tabular}   & \begin{tabular}[c]{@{}c@{}}49.14\\ (44\%)\end{tabular}   \\ \hline
LVI-SAM + Zhang et al. \cite{Zhang2016_zhang_etal} & \begin{tabular}[c]{@{}c@{}}23.26\\ (100\%)\end{tabular}  & \begin{tabular}[c]{@{}c@{}}16.91\\ (100\%)\end{tabular}  & \begin{tabular}[c]{@{}c@{}}0.44\\ (100\%)\end{tabular}  & \begin{tabular}[c]{@{}c@{}}0.74\\ (100\%)\end{tabular}   & \begin{tabular}[c]{@{}c@{}}1.27\\ (100\%)\end{tabular}  & \begin{tabular}[c]{@{}c@{}}3.32\\ (100\%)\end{tabular}   & \begin{tabular}[c]{@{}c@{}}1.43\\ (100\%)\end{tabular}   & \begin{tabular}[c]{@{}c@{}}12.10\\ (100\%)\end{tabular}  & \begin{tabular}[c]{@{}c@{}}4.39\\ (100\%)\end{tabular}   & \begin{tabular}[c]{@{}c@{}}36.60\\ (100\%)\end{tabular}  \\ \hline
LVI-SAM + Xn-ICP \cite{Tuna2024_xicp}       & \begin{tabular}[c]{@{}c@{}}8.65\\ (100\%)\end{tabular}  & \begin{tabular}[c]{@{}c@{}}18.08\\ (100\%)\end{tabular}  & \begin{tabular}[c]{@{}c@{}}0.40\\ (100\%)\end{tabular} & \begin{tabular}[c]{@{}c@{}}0.74\\ (100\%)\end{tabular}   & \begin{tabular}[c]{@{}c@{}}1.11\\ (100\%)\end{tabular}  & \begin{tabular}[c]{@{}c@{}}3.07\\ (100\%)\end{tabular}   & \begin{tabular}[c]{@{}c@{}}1.43\\ (100\%)\end{tabular}   & \begin{tabular}[c]{@{}c@{}}\textbf{3.85}\\ (100\%)\end{tabular}   & \begin{tabular}[c]{@{}c@{}}5.29\\ (100\%)\end{tabular}   & \begin{tabular}[c]{@{}c@{}}36.75\\ (100\%)\end{tabular}  \\ \hline
LVI-SAM + Ours         & \begin{tabular}[c]{@{}c@{}}\textbf{6.26}\\ (100\%)\end{tabular}   & \begin{tabular}[c]{@{}c@{}}\textbf{7.44}\\ (100\%)\end{tabular}   & \begin{tabular}[c]{@{}c@{}}\textbf{0.38}\\ (100\%)\end{tabular}  & \begin{tabular}[c]{@{}c@{}}\textbf{0.43}\\ (100\%)\end{tabular}   & \begin{tabular}[c]{@{}c@{}}\textbf{0.48}\\ (100\%)\end{tabular}  & \begin{tabular}[c]{@{}c@{}}\textbf{1.56}\\ (100\%)\end{tabular}   & \begin{tabular}[c]{@{}c@{}}\textbf{1.40}\\ (100\%)\end{tabular}   & \begin{tabular}[c]{@{}c@{}}5.66\\ (100\%)\end{tabular}   & \begin{tabular}[c]{@{}c@{}}\textbf{2.33}\\ (100\%)\end{tabular}   & \begin{tabular}[c]{@{}c@{}}\textbf{12.31}\\ (100\%)\end{tabular} \\ \hline
\end{tabular}
\begin{tablenotes}[para,flushleft]
        \item The units are in meters. Bold values indicate the best results. 
        \item``\%" denotes the length of the ground truth associated with the estimated trajectory divided by the total length of the ground truth for the sequence.
     \end{tablenotes} 
\end{threeparttable} 
\end{table*}

\subsection{Simulation Experiments in a Planetary-like Environment}
\label{section:experiment_B}
Planetary-like environments are typical degraded scenarios that pose significant challenges to existing SLAM algorithms. Conducting field experiments and collecting data directly in unstructured environments such as the Moon or Mars is currently difficult and costly. Therefore, we collected a simulated SLAM dataset, the PLAM dataset, in a visually realistic lunar-like simulation environment, which includes sensor data from LiDAR, a camera, and an IMU.

\subsubsection{PLAM Dataset}
\begin{figure}[t]
\includegraphics[ width=\linewidth]{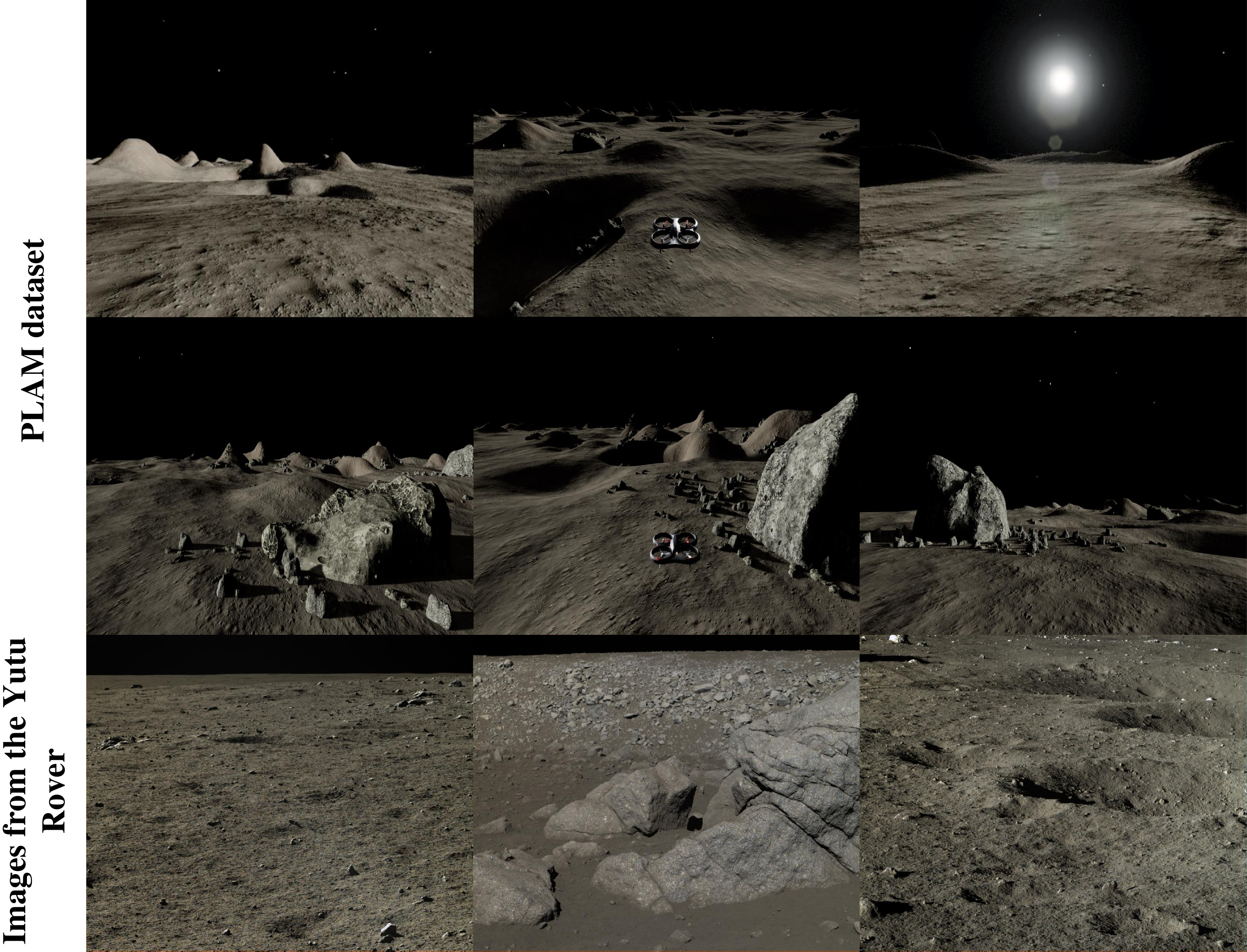}
\caption{Simulated environment of the PLAM dataset and lunar color images from the Yutu rover are shown. The lunar images are sourced from [https://planetary.s3.amazonaws.com/data/change3/pcam.html]. 
}
\label{fig:environment_exp1}
\vspace{0 em}
\end{figure}

We built a visually realistic lunar-like simulation environment using AIRSIM and UE4. The scene's visual rendering is based on the color photographs of the Moon taken by the Yutu rover, as shown in Fig.~\ref{fig:environment_exp1} for comparison. A drone equipped with a monocular camera, IMU, and LiDAR was used to collect the PLAM dataset, which is designed to test and evaluate the localization accuracy of SLAM algorithms in large-scale extreme environments. The dataset is provided in rosbag format and includes the ground truth trajectory of the robot during its operation.

The scene features uneven terrain with hills, pits, as well as rocky and non-rocky areas. It also presents challenges due to varying lighting conditions. The PLAM dataset exhibits typical characteristics of planetary-like environments, such as visual ambiguity, blurriness, a lack of features and geometric structure, as well as variations in lighting and shadowed regions. The PLAM dataset includes 10 sequences, with a total trajectory length of 10 km. It can be used to test different SLAM modules and evaluate the performance of various tasks in planetary exploration, such as odometry, loop closure detection, home and traverse tasks, and more. The rocky areas contain rocks of various sizes, providing more visual features and geometric structure. Non-rocky areas pose a greater challenge for both vision-based and LiDAR-based SLAM methods.

\begin{figure*}[ht]
\centering
\includegraphics[width=1.0\linewidth]{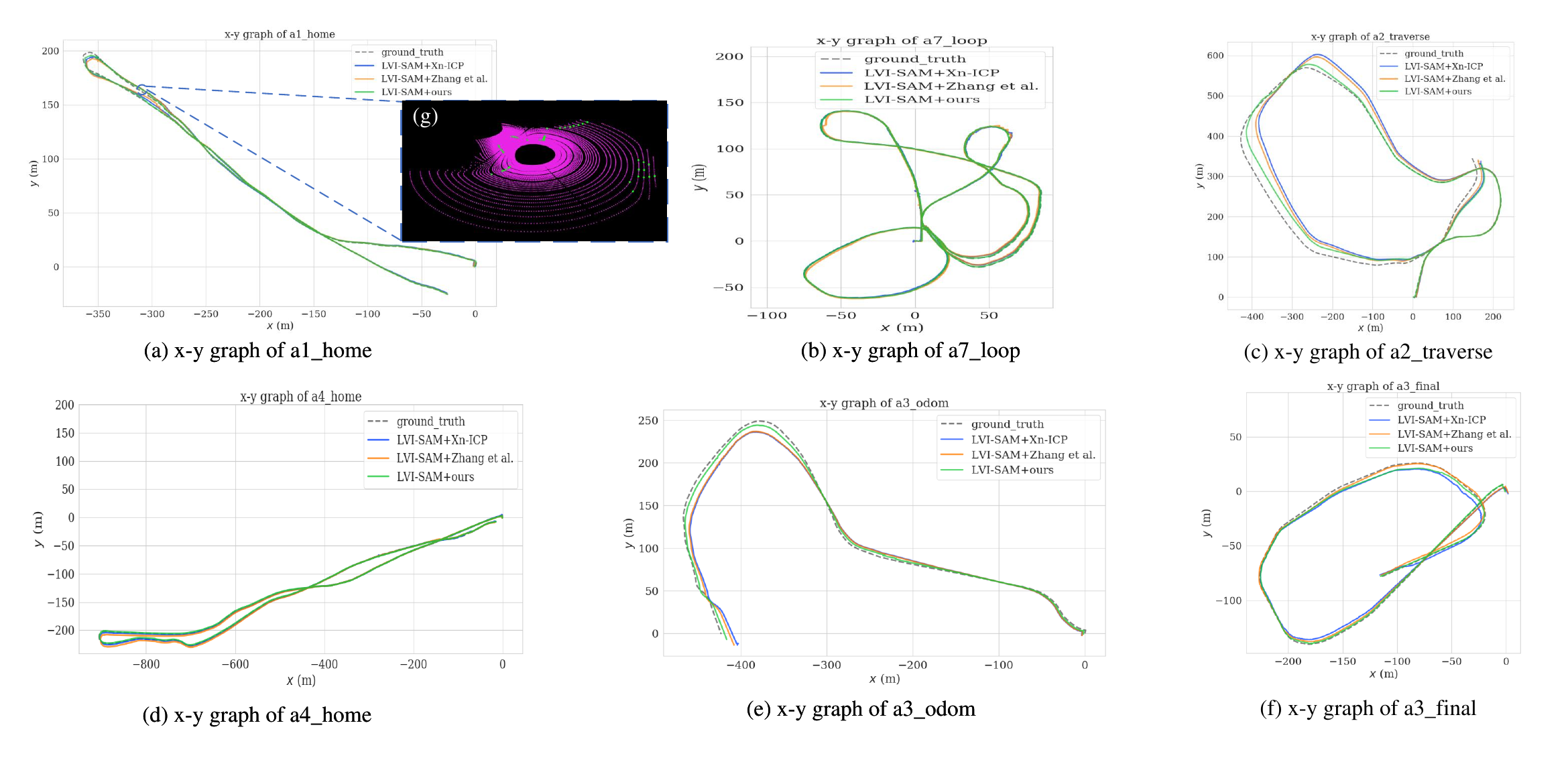}
\caption{(a)-(f) Trajectories estimated by LVI-SAM + Ours and the state-of-the-art methods on the sequences a1\_home, a3\_odom, a2\_traverse, a4\_home, a7\_loop, and a3\_final, and ground truth trajectories are shown. (g) Edge points (green) and planar points (purple) extracted by the front end of LVI-SAM from a LiDAR scan during the a1\_home sequence.}
\label{fig:exp1result}
\end{figure*}

\begin{figure}[ht]
\includegraphics[ width=\linewidth, height=10cm]{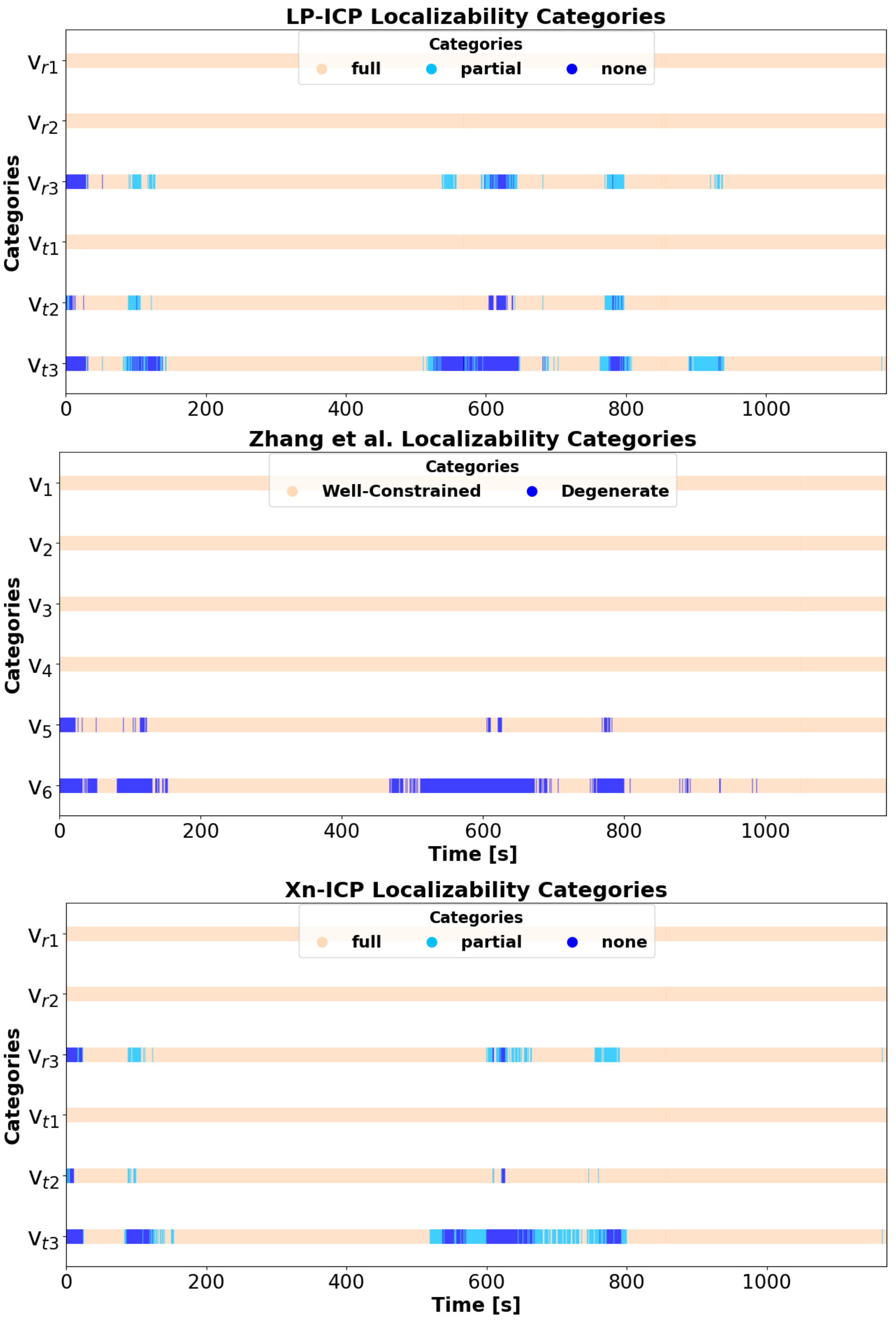}
\caption{Estimated localizability categories of LP-ICP and the state-of-the-art methods in the a2\_traverse sequence.}
\label{fig:localizability_exp1}
\vspace{0 em}
\end{figure}

\subsubsection{Results}
We evaluate the localization performance of LVI-SAM + Ours, LVI-SAM + Zhang et al. \cite{Zhang2016_zhang_etal}, and LVI-SAM + Xn-ICP \cite{Tuna2024_xicp} on the PLAM dataset and compare and analyze the results. At the same time, we test and evaluate the performance of some advanced SLAM systems, such as ORB-SLAM3 \cite{Campos2021_orbslam3} in monocular camera-IMU mode, VINS-MONO \cite{Qin2018_vinsmono}, and LIO-SAM \cite{Shan2020_liosam}. The estimated trajectory obtained by each algorithm with the ground truth trajectory are compared. The first 50 poses of both trajectories are aligned using the EVO tool \cite{Grupp2018evo}, and the Absolute Trajectory Error (ATE) \cite{Sturm2012_ate} between them is calculated. The results are presented in the form of root mean square error (RMSE). Additionally, we provide an extra metric, which is the ratio of the algorithm’s estimated poses to the total length of the ground truth for each sequence. This value is calculated as the length of the ground truth associated with the estimated trajectory divided by the total length of the ground truth for the sequence. When large drift or fluctuations in the trajectory occurred, preventing the algorithm from continuing, we stopped the experiment. We compare the accuracy among algorithms that successfully complete the entire sequence. For those that did not complete the sequence, we provided their ATE to better understand their performance on the specific sequence. Table~\ref{table:exp1result} reports these metrics for the tested algorithms on each sequence of the dataset. These results show that the extreme environment of the PLAM dataset poses significant challenges to VINS-MONO, ORB-SLAM3, and LIO-SAM. The long-term visual ambiguity, lack of features and geometric structure, and poor lighting conditions in the PLAM dataset lead to instability and low accuracy during their operation. In comparison, the multi-sensor fusion algorithm LVI-SAM achieves better accuracy than them. 

The degeneracy detection algorithms in LVI-SAM + Ours, LVI-SAM + Zhang et al. \cite{Zhang2016_zhang_etal}, and LVI-SAM + Xn-ICP \cite{Tuna2024_xicp} were further compared. The trajectories estimated by these three methods on six sequences of the dataset are presented in Fig.~\ref{fig:exp1result}. Fig.~\ref{fig:exp1result} (g) shows the edge points and planar points extracted from a LiDAR scan during the a1\_home sequence, when degeneracy is detected along a translational direction in the x-y plane. Our method, LP-ICP, achieves better localization accuracy on multiple sequences. LVI-SAM + Zhang et al. \cite{Zhang2016_zhang_etal} also demonstrates good accuracy in the rocky-area sequences a2\_home, a4\_loop, and a7\_loop, as the abundance of rocks provides more features and geometric structures. However, in sequences from other non-rocky areas, on the one hand, the method cannot pick out high-contribution correspondence constraints in ill-conditioned directions. On the other hand, the scales of rotational and translational directions differ. If the same threshold settings are applied, it limits the degeneracy detection, leading to a decrease in performance. For instance, as shown in Fig.~\ref{fig:localizability_exp1}, in the a2\_traverse sequence, when the robot is in an open area of the scene at 620s, LiDAR degeneracy occurs in the three directions: translation along the x-axis and y-axis, and rotation around the z-axis. However, the method by Zhang et al. \cite{Zhang2016_zhang_etal} can only detect two of these directions. X-ICP improves performance by adding partial localizability categories and unifying the scale of localizability for both rotation and translation.  In multiple sequences such as a2\_odom and a4\_home, it shows better accuracy compared to Zhang et al. \cite{Zhang2016_zhang_etal}. However, X-ICP cannot utilize the localizability information from edge point-to-line correspondences. As shown in Fig.~\ref{fig:localizability_exp1}, our method considers more localizability information to be usable. Additionally, in partially localizable directions, the added soft constraints enable pose updates to be performed under additional constraints and help improve accuracy. However, in our experiments, we observed that the use of soft constraints in partially localizable directions may introduce fluctuations and uncertainty in pose estimation. The underlying causes and potential improvements will be investigated in future work.

\begin{table*}[ht]
\centering
\caption{The RMSE of ATE and Ratio of Completion of the Compared Algorithms on the CERBERUS DARPA Subterranean Challenge Dataset}
\label{table:exp2result}
\begin{threeparttable}  
\renewcommand{\arraystretch}{1.2} 
\begin{tabular}{ccccc}
\hline
\rule[-1.5ex]{0pt}{4ex} 
                       & ANYmal 1     & ANYmal 2      & ANYmal 3      & ANYmal 4      \\ \hline
VINS-Mono \cite{Qin2018_vinsmono}              & 2.56 (100\%) & 2.17 (100\%)  & 3.94 (100\%)  & 10.43 (100\%) \\
LIO-SAM \cite{Shan2020_liosam}               & 8.91 (100\%) & 20.89 (100\%) & 18.51 (100\%) & 11.20 (100\%) \\
R3LIVE \cite{Lin2022_r3live}                & 1.29 (20\%)  & 8.11 (12\%)   & 2.06 (100\%)  & 14.00 (100\%)  \\
LVI-SAM + Zhang et al. \cite{Zhang2016_zhang_etal} & 0.46 (100\%) & 0.44 (100\%)  & 0.50 (100\%)  & \textbf{0.37} (100\%)  \\
LVI-SAM + Xn-ICP \cite{Tuna2024_xicp}        & \textbf{0.30} (100\%) & 0.32 (100\%)  & 0.56 (100\%)  & 0.43 (100\%)  \\
LVI-SAM + Ours         & 0.37 (100\%) & \textbf{0.24} (100\%)  & \textbf{0.37} (100\%)  & 0.39 (100\%)  \\ \hline
\end{tabular}
\begin{tablenotes}[para,flushleft]
        \item The units are in meters. Bold values indicate the best results. 
        \item ``\%" denotes the length of the ground truth associated with the estimated trajectory divided by the total length of the ground truth for the sequence.
     \end{tablenotes} 
\end{threeparttable} 
\end{table*}

\subsection{Real-World Experiment on the CERBERUS DARPA Subterranean Challenge Dataset}
\label{section:experiment_C}
The CERBERUS DARPA Subterranean Challenge Datasets \cite{Tranzatto2024_cerberus} were collected in the Louisville Mega Cavern in Kentucky. During the Final Event of the DARPA Subterranean (SubT) Challenge, the quadruped robot ANYmal, equipped with cameras, IMU, and LiDAR, collected the CERBERUS dataset. The underground tunnel environment is shown in Fig.~\ref{fig:intro_environment}. Features such as poor lighting conditions, cave environments resembling long corridors, self-similar areas, and motion vibrations caused by rugged terrain pose challenges to existing SLAM methods \cite{Ebadi2024_darpa}. These factors result in increased localization and mapping errors. The dataset contains four sequences.

We test and evaluate the localization and mapping performances of LVI-SAM + Ours, LVI-SAM + Xn-ICP \cite{Tuna2024_xicp}, and LVI-SAM + Zhang et al. \cite{Zhang2016_zhang_etal} on the CERBERUS dataset. Additionally, the performances of several state-of-the-art SLAM systems, such as VINS-MONO, LIO-SAM, and R3LIVE \cite{Lin2022_r3live}, is also tested. Similarly, we use the EVO tool to evaluate the results. The estimated trajectory is aligned with the ground truth trajectory and the Absolute Trajectory Error (ATE) is calculated. Table~\ref{table:exp2result} reports the localization performance metrics of the tested algorithms on the various sequences of the dataset. Note that the results of the R3LIVE are presented using the better performance between our test results and those from \cite{Lee2024_switchslam}.

\begin{figure}[t]
\includegraphics[ width=\linewidth, height=10cm]{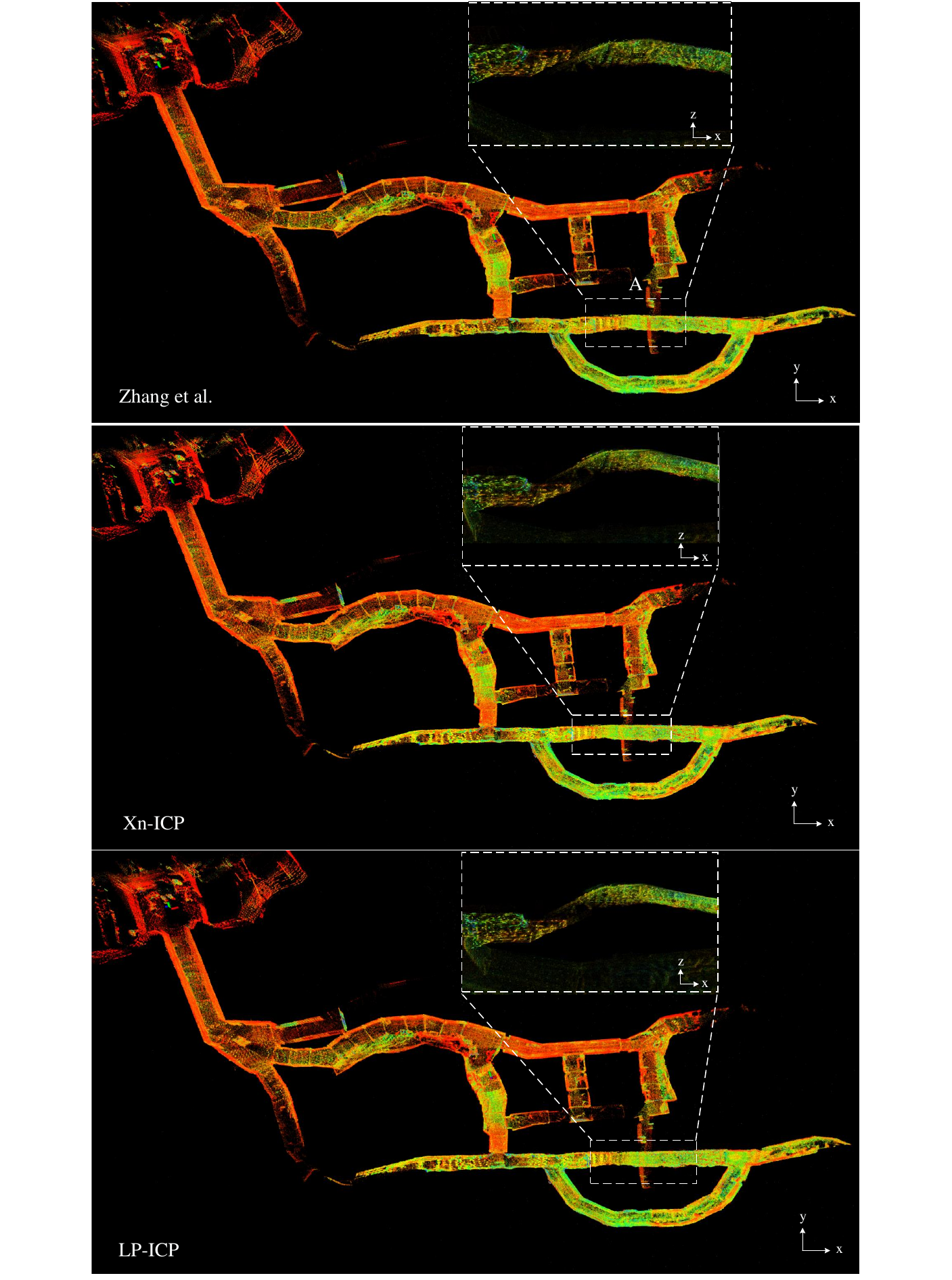}
\caption{Resulting maps from the compared methods and LP-ICP on the ANYmal 1 sequence.}
\label{fig:exp2result_map}
\vspace{0 em}
\end{figure}

As shown in Table~\ref{table:exp2result}, the characteristics of underground tunnel pose significant challenges to many methods. Poor lighting conditions and darkness, along with self-similar scenes, create difficulties for vision-based approaches. LiDAR -based methods struggle with accuracy in degenerate scenarios, such as long corridor-like passages in tunnels. Uneven terrain and vibrations during movement also negatively impact the final localization results. These factors result in unsatisfactory accuracy for  VINS-MONO, LIO-SAM, and R3LIVE. By employing multi-sensor fusion and utilizing the VIO submodule to provide a relatively reliable initial estimate, along with incorporating degeneracy detection and handling, LVI-SAM + Ours, LVI-SAM + Zhang et al., and LVI-SAM + Xn-ICP achieve significantly better localization accuracy. Among these three localizability-aware algorithms, our method achieves better accuracy on the ANYmal 2 and ANYmal 3 sequences. On the ANYmal 1 and ANYmal 4 sequences, its accuracy is comparable to that of LVI-SAM + Xn-ICP and LVI-SAM + Zhang et al.

Additionally, the mapping results of the three methods on the ANYmal 1 sequence are shown in Fig.~\ref{fig:exp2result_map}. In the dashed box of Fig.~\ref{fig:exp2result_map}, which marks the tunnel connection area A, the robot approached A from both ends during operation but ultimately did not reach this location. This posed a challenge for mapping. As shown in the mapping result of Zhang et al. \cite{Zhang2016_zhang_etal}, which presents the front view of location A in the x-z plane, the mapping result contains an error where the top of the tunnel on the right side of A is lower than the ground of the left side tunnel. This causes an incorrect tunnel connection at location A. In contrast, by using fine-grained degeneracy detection methods, the results of LVI-SAM + Ours and LVI-SAM + Xn-ICP do not exhibit this error.

\begin{figure*}[t]
\centering
\includegraphics[ width=\linewidth, height=10cm]{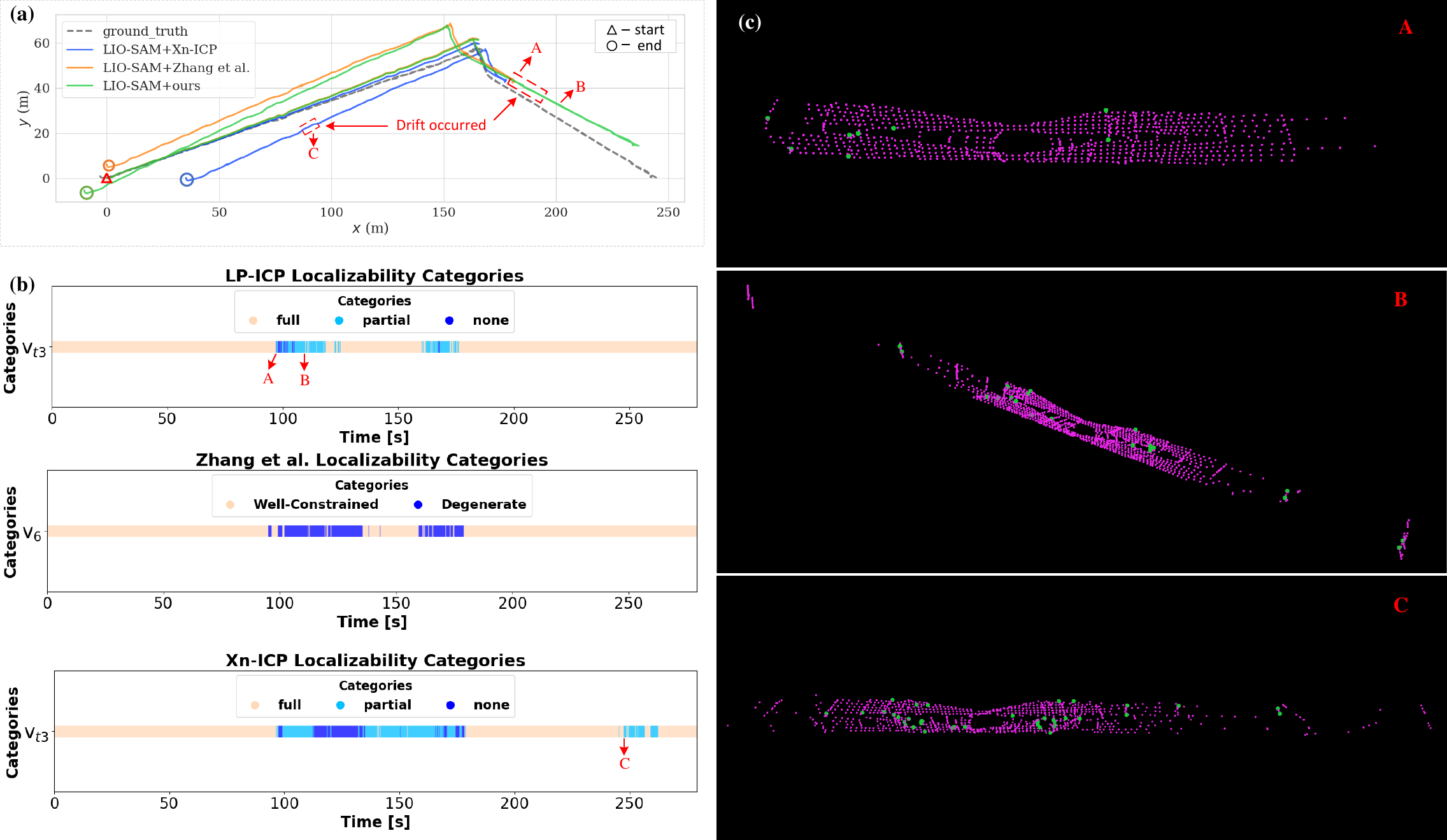}
\caption{(a) Trajectories estimated by three methods and ground truth trajectory are shown.  (b) Estimated localizability categories of LP-ICP and the state-of-the-art methods. (c) Edge points (green) and planar points (purple) extracted from three LiDAR scans collected at locations A, B, and C in (a), respectively.}
\label{fig:exp3result_map}
\vspace{0 em}
\end{figure*}

\subsection{SubT-MRS Dataset}
\label{section:experiment_subt}
The SubT-MRS Dataset \cite{Zhao2024_subtdataset} is a challenging real-world dataset collected from subterranean environments, including caves, urban areas, and tunnels. It was used in the ICCV 2023 SLAM Challenge. The dataset introduces challenging conditions such as textureless surfaces, smoke, illumination changes, and featureless structures.

We conducted tests on the Long Corridor sequence from the SubT-MRS dataset. In this scenario, the localization accuracy of the ground robot is primarily affected by degeneracy along the longitudinal translational direction of the corridor. In this experiment, camera images were not used, and loop closure detection was disabled. Since LVI-SAM becomes similar to LIO-SAM when image topics are not subscribed to, the three compared methods are referred to as LIO-SAM + Zhang et al., LIO-SAM + Xn-ICP, and LIO-SAM + Ours. 

Table~\ref{table:experiment_subt_result} presents the localization accuracy of the three compared methods. The estimated trajectories of the three methods are shown in Fig.~\ref{fig:exp3result_map} (a). The localization errors are primarily caused by drift at two specific locations. Fig.~\ref{fig:exp3result_map} (b) shows the degeneracy detection results of the three methods along the translational direction with the weakest constraints. As shown in Fig. 9(a), all three methods begin to exhibit noticeable drift after reaching position A from the start point. This is due to the lack of sufficient information from edge or planar features along the corridor direction, as shown in Fig.~\ref{fig:exp3result_map} (c) A. Our method incorporates additional point-to-line localizability information near location B, resulting in reduced drift. Note that between 130s and 170s, when the robot approaches the end of the corridor, LIO-SAM + Xn-ICP detects degeneracy along the translational direction. We attribute this to drift-induced mapping errors, which result in incorrect scan-to-map matching and erroneous normal vector estimation in the front end. This issue is not inherent to X-ICP itself. LIO-SAM + Xn-ICP exhibits drift near position C. In contrast, LIO-SAM + Zhang et al. and LIO-SAM + Ours utilize edge point-to-line geometric constraints at this location, as shown in Fig.~\ref{fig:exp3result_map} (c) C.

\begin{table}[t]
\centering
\caption{The RMSE of ATE of the Compared Algorithms on the Long Corridor sequence from the SubT-MRS dataset}
\label{table:experiment_subt_result}
\begin{threeparttable}
\renewcommand{\arraystretch}{1.2}
\begin{tabular}{ccccc}
\hline
\rule[-1.5ex]{0pt}{4ex}   & ATE [m] \\
\hline
LIO-SAM + Zhang et al. \cite{Zhang2016_zhang_etal} & 24.71 \\
LIO-SAM + Xn-ICP \cite{Tuna2024_xicp}              & 19.63 \\
LIO-SAM + Ours                                     & \textbf{11.92} \\
\hline
\end{tabular}
\begin{tablenotes}[para,flushleft]
\item Bold value indicates the best result.
\end{tablenotes}
\end{threeparttable}
\end{table}

\subsection{Runtime Evaluation}
\label{section:experiment_runtime}
Table~\ref{table:runtime_result} shows the runtime results of LP-ICP on the ANYmal 1 sequence of the CERBERUS DARPA subterranean challenge dataset. We tested the average time consumption of the scan-to-map registration module for three methods: LVI-SAM (without degeneracy detection), LVI-SAM + Ours, and LVI-SAM + Zhang et al \cite{Zhang2016_zhang_etal}. The tests were conducted on a laptop with an Intel i7-12700H CPU. As shown in Table~\ref{table:runtime_result}, the time consumption of the three methods is quite similar. Interestingly, the method without degeneracy detection takes the most time, which we believe is due to optimization in ill-conditioned directions that makes it harder to converge, resulting in an increased number of iterations. This demonstrates that our method, LP-ICP, has the capability to run in real time on robotic systems.

\begin{table}[t]
\centering
\caption{The Average Time Consumption (Milliseconds) of Scan-to-Map Registration Using Different Methods on the ANYmal 1 Sequence}
\label{table:runtime_result}
\begin{threeparttable}
\renewcommand{\arraystretch}{1.2}
\begin{tabular}{ccccc}
\hline
\rule[-1.5ex]{0pt}{4ex}  & Time Consumption\\& (Average) \\
\hline
LVI-SAM + Ours                                 & 35.87 \\
LVI-SAM + Zhang et al. \cite{Zhang2016_zhang_etal} & 33.69 \\
LVI-SAM (without Degeneracy Detection)         & 38.04 \\
\hline
\end{tabular}
\end{threeparttable}
\end{table}

\section{Conclusion}
To improve the accuracy of LiDAR-based SLAM algorithms in challenging unstructured environments, degeneracy detection and handling are essential. This paper proposes a new ICP algorithm framework, LP-ICP, which combines point-to-line and point-to-plane cost functions, along with localizability detection and handling. LP-ICP detects degeneracy by calculating the localizability contribution of geometric correspondences and adds additional constraints to improve the accuracy of pose estimation. The efficacy of the proposed method is validated through experimental evaluation on our planetary-like simulation dataset and two real-world datasets, demonstrating comparable or improved performance over state-of-the-art methods under our tested settings. Currently, the performance of degeneracy detection and optimization algorithms is influenced by their parameter settings. In our future work, we will validate the sensitivity of the algorithm parameters to different environments and hardware configurations, as well as investigate the impact of performing localizability analysis separately in the translational and rotational eigenspaces. In addition, the partial localizable category with soft constraints introduces trajectory jitter and uncertainty in the experimental results. In future work, we will further investigate the underlying causes and explore potential improvements.

\bibliographystyle{IEEEtran}
\bibliography{myreference} 

\vspace{-1.5cm}

\vspace{-1.5cm}

\newpage

\vfill

\end{document}